%% file: main.tex
\documentclass[runningheads]{llncs}
\usepackage[T1]{fontenc}
\usepackage{graphicx}
\usepackage{booktabs}
\usepackage[misc]{ifsym}
\usepackage{amssymb}

\usepackage[numbers]{natbib}
\usepackage{float}
\usepackage{placeins}
\usepackage{tabularx}
\usepackage{makecell}
\usepackage{rotating}

\usepackage{subcaption}
\usepackage{algorithm}
\usepackage{algpseudocode}
\usepackage[most]{tcolorbox}

\definecolor{darkgreen}{RGB}{0,110,0}

\definecolor{colL1}{RGB}{95, 45, 125}   % Plum / Academic Purple
\definecolor{colL1bg}{RGB}{247, 244, 250} % Lavender background
\definecolor{colL2}{RGB}{185, 90, 0}    % Burnt Orange
\definecolor{colL2bg}{RGB}{252, 248, 242}
\definecolor{colL3}{RGB}{0, 100, 80}    % Deep Teal
\definecolor{colL3bg}{RGB}{242, 250, 248}

\usepackage{colortbl}
\usepackage{adjustbox}
\usepackage{multirow}
\usepackage{array}
\usepackage[table]{xcolor}
\usepackage{mdframed}
\usepackage{ragged2e}

\usepackage{pgf}
\usepackage{collcell}

\definecolor{roadretrievalcolor}{HTML}{7FA8D1} % Deep Steel Blue
\definecolor{spatialcalccolor}{HTML}{B3A369}   % Rich Taupe / Sand
\definecolor{metadatapredcolor}{HTML}{A77EB3}  % Royal Lavender

\newcommand{\colorizehigh}[2]{%
  \if\relax\detokenize{#2}\relax\else% Check for empty cell
    \pgfmathsetmacro{\temppercent}{int(max(0, min(100, #2*100)))}%
    \edef\temp{\noexpand\cellcolor{#1!\temppercent!white}}%
    \temp
    #2%
  \fi%
}

\newcommand{\colorizelow}[2]{%
  \if\relax\detokenize{#2}\relax\else% Check for empty cell
    \pgfmathsetmacro{\temppercent}{int(max(0, min(100, (#2 > 1.5) ? 0 : (1 - #2/1.5)*100)))}%
    \edef\temp{\noexpand\cellcolor{#1!\temppercent!white}}%
    \temp
    #2%
  \fi%
}

\newcolumntype{H}{>{\collectcell{\colorizehigh{roadretrievalcolor}}}c<{\endcollectcell}}

\newcolumntype{I}{>{\collectcell{\colorizehigh{spatialcalccolor}}}c<{\endcollectcell}} % Higher is better
\newcolumntype{K}{>{\collectcell{\colorizelow{spatialcalccolor}}}c<{\endcollectcell}}   % Lower is better

\newcolumntype{J}{>{\collectcell{\colorizehigh{metadatapredcolor}}}c<{\endcollectcell}} % Higher is better
\newcolumntype{M}{>{\collectcell{\colorizelow{metadatapredcolor}}}c<{\endcollectcell}}   % Lower is better

\newcolumntype{L}[1]{>{\RaggedRight\arraybackslash}p{#1}}
\newcolumntype{C}[1]{>{\Centering\arraybackslash}p{#1}}
\newcolumntype{Y}{>{\RaggedRight\arraybackslash}X}

\AtBeginDocument{%
  }

\begin{document}

\title{GeoResponder: Towards Building Geospatial LLMs for Time-Critical Disaster Response}
\titlerunning{GeoResponder}

% \author{Author information scrubbed for double-blind reviewing}

\author{Ahmed El Fekih Zguir\inst{1} \and Ferda Ofli\inst{1} \and Muhammad Imran\inst{1}}
\authorrunning{A.E.F. Zguir et al.}

\institute{Qatar Computing Research Institute, Doha, Qatar \email{\{azguir, fofli, mimran\}@hbku.edu.qa}}

\maketitle

\begin{abstract}
LLMs excel at linguistic tasks but lack the inner geospatial capabilities needed for time-critical disaster response, where reasoning about road networks, coordinates, and access to essential infrastructure such as hospitals, shelters, and pharmacies is vital. We introduce GeoResponder, a framework that instills robust spatial reasoning through a scaffolded instruction-tuning curriculum. By stratifying geospatial learning into different cognitive layers, we anchor semantic knowledge to the continuous coordinate manifold and enforce the internalization of spatial axioms. Extensive evaluations across four topologically distinct cities and diverse tasks demonstrate that GeoResponder significantly outperforms both state-of-the-art foundation models and domain-specific baselines. These results suggest that LLMs can begin to internalize and generalize geospatial structures, pointing toward the future development of language models capable of supporting disaster response needs.
\keywords{Geospatial reasoning \and Disaster response \and LLMs}
\end{abstract}

% ------------ Figure 7: Qualitative comparison (3 models + prompt)
\begin{figure}[t]
\centering

% --- instruction strip ---
% \begin{tcolorbox}[
%     width=\textwidth,
%     colback=gray!5,
%     colframe=gray!40,
%     boxrule=0.5pt,
%     arc=2pt,
%     left=6pt,right=6pt,top=4pt,bottom=4pt
% ]
\small
\textbf{\textcolor{blue!60!black}{Task:}} We have an injured person at the coordinates \textcolor{teal!60!black}{$(48.874253,\ 2.348557)$}. Which hospital is closest?
\hfill
% \end{tcolorbox}

\vspace{8pt}

% --- model outputs ---
\begin{subfigure}[t]{0.33\textwidth}
    \centering
    \includegraphics[width=\linewidth]{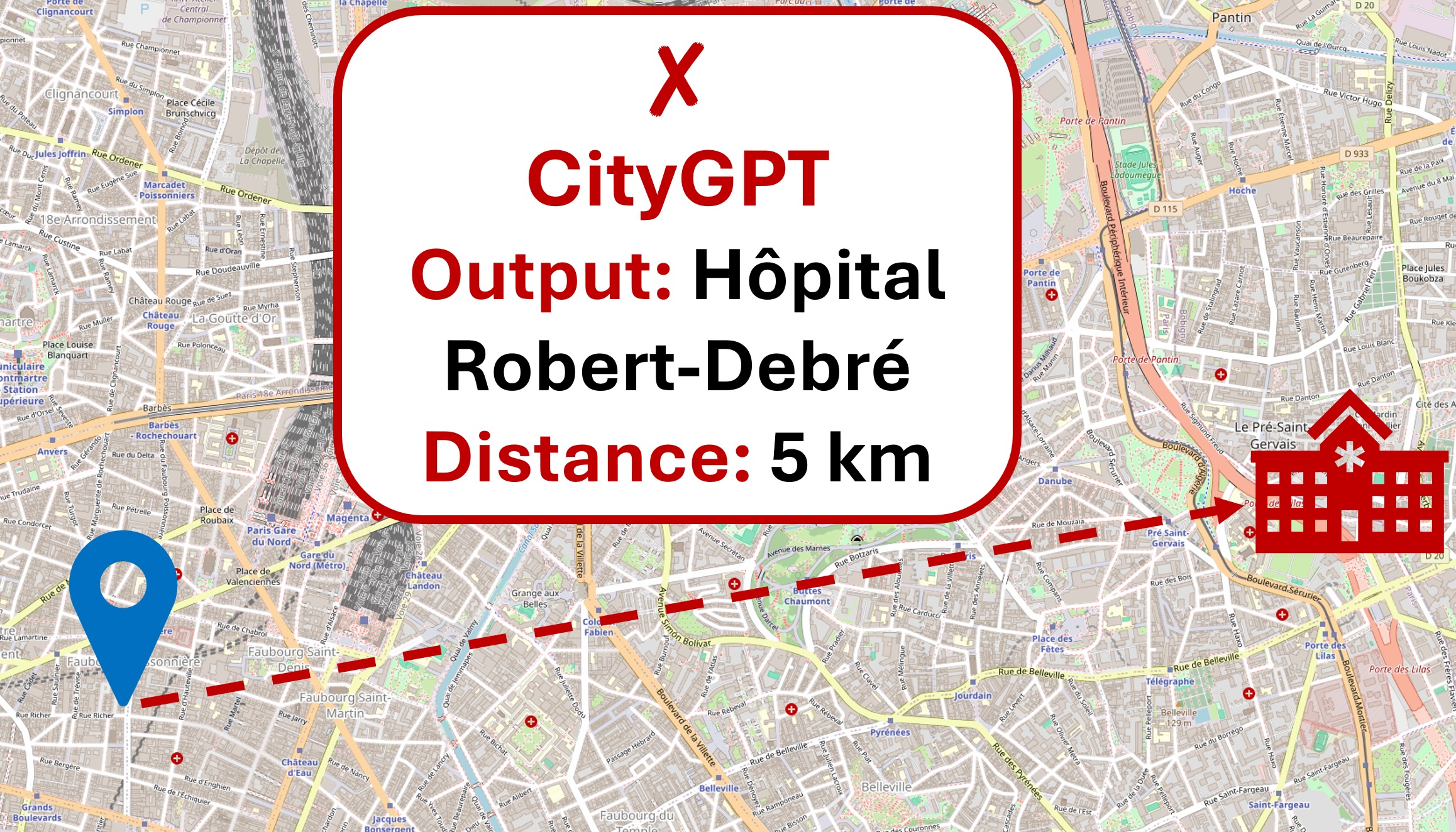}
    \caption{CityGPT}
\end{subfigure}\hfill
\begin{subfigure}[t]{0.33\textwidth}
    \centering
    \includegraphics[width=\linewidth]{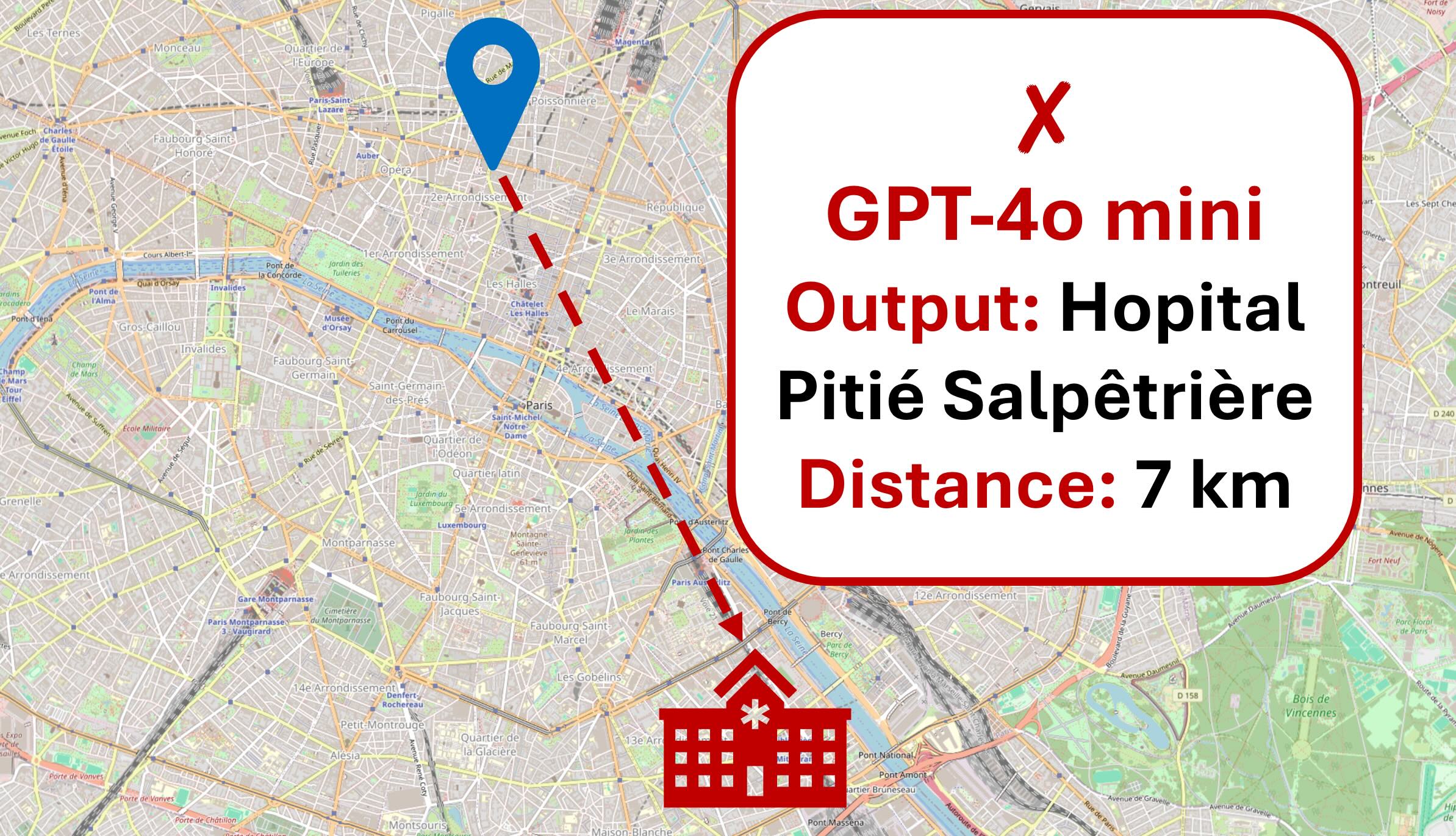}
    \caption{GPT-4o Mini}
\end{subfigure}\hfill
\begin{subfigure}[t]{0.33\textwidth}
    \centering
    \includegraphics[width=\linewidth]{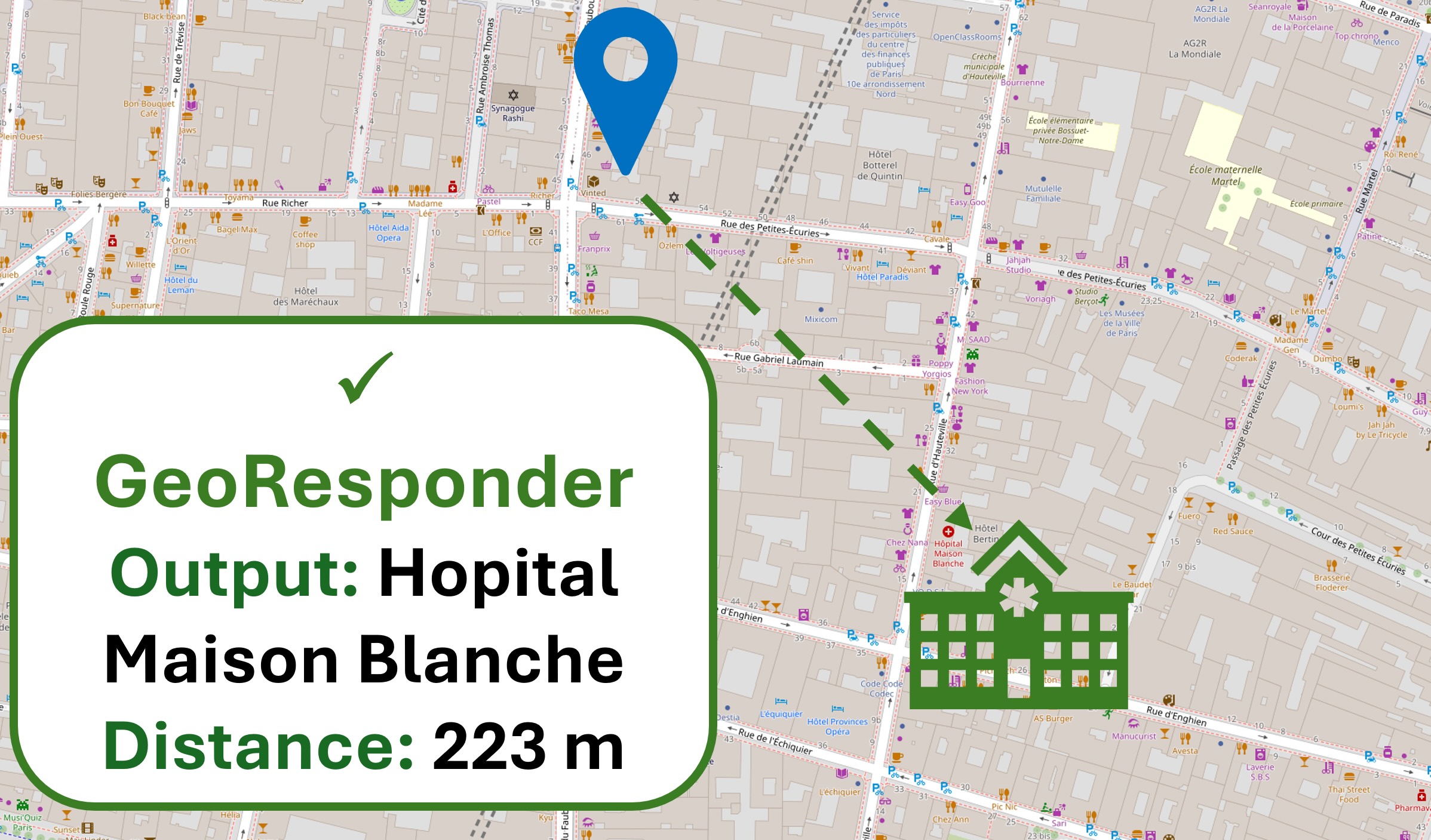}
    \caption{GeoResponder (Ours)}
\end{subfigure}

\caption{Qualitative comparison of baseline models and GeoResponder.}
\label{fig:qualitative_hospital}

\end{figure}

% main parts of the paper
\input{introduction}

\input{related_work}

\input{methodology}
\input{datasets}
\input{evaluation_and_results}

\input{ablation}
\input{conclusion}

\begingroup
\medskip
\let\clearpage\relax
\bibliographystyle{splncs04}
\bibliography{references}
\endgroup

%% Appendix
% \appendix
% \input{appendix}

\end{document}

%% file: introduction.tex
\section{Introduction}

% In the chaotic aftermath of a natural disaster, the efficacy of emergency response depends on the velocity of decision-making. First responders and on-ground coordinators are confronted with a deluge of critical inquiries, e.g., identifying the nearest accessible schools for shelter, verifying road network connectivity for supply convoys, or locating hospitals within a serviceable radius~\cite{nearing:24,singh:25, kolivand:24, skai:24}. While the geospatial data necessary to answer these questions exists, often meticulously detailed in platforms like OpenStreetMap (OSM), it remains effectively locked behind complex Geographic Information Systems (GIS) requiring specialized expertise. This disconnect creates a dangerous ``accessibility gap'': at the very moment when data fluidity is most vital, the cognitive load of translating urgent natural language intent into rigid spatial database queries becomes a bottleneck that compromises public safety.

In the chaotic aftermath of a natural disaster, the efficacy of emergency response depends on the velocity of decision-making. First responders are confronted with a deluge of critical inquiries, e.g., identifying the nearest accessible schools for shelter, verifying road connectivity for supply convoys, or locating hospitals within a serviceable radius~\cite{nearing:24,singh:25, skai:24}. Although geospatial data for these tasks exists in platforms like OpenStreetMap (OSM), it remains locked behind complex Geographic Information Systems (GIS) requiring specialized expertise. This disconnect creates a dangerous ``accessibility gap'': when data fluidity is most vital, the cognitive burden of translating natural language intent into rigid spatial queries becomes a bottleneck that compromises public safety.

Large Language Models (LLMs) offer a promising bridge across this gap by enabling responders to pose complex queries in natural language. However, while general-purpose LLMs possess high semantic fluency, they exhibit a profound deficit in geospatial reasoning (Figure~\ref{fig:qualitative_hospital}). Recent ``agentic'' frameworks attempt to mitigate this by outsourcing spatial queries to external tools. While such agents remain indispensable for retrieving transient, real-time states, they are fundamentally brittle when tasked with complex spatial reasoning. Consider the query: ``Find the nearest clinic to the shelter accessible without crossing the river.'' A standard agent, relying on a general LLM to formulate tool parameters, will often fail to recognize the river as a topological barrier, retrieving a facility that is geometrically close but physically inaccessible. Although general LLMs may memorize that a specific hospital is in a specific city, they cannot reliably infer topology (how roads connect), metrics (the precise distance between two shelters), or orientation (the cardinal direction of an evacuation route)~\cite{Ji:25, cohn:24, bhandari:23}.

To address these limitations, we propose a shift from latent textual correlation to structured geospatial supervision. Rather than inferring proximity because two locations appear together in text, our approach teaches the model to verify their relationship through explicit coordinate-based rules. We introduce GeoResponder, a framework that translates deterministic GIS operations into natural language training signals, systematically bridging the gap between map data and language. Specifically, our methodology stratifies geospatial intelligence into three cognitive layers: \textbf{(i) Spatial grounding}, which anchors the model in static knowledge (e.g., resolving the entity ``City Hospital'' to its precise coordinates); \textbf{(ii) Spatial reasoning}, which enforces consistent logic for physical properties (e.g., calculating Haversine distance or verifying road intersections); and \textbf{(iii) Constraints-aware Spatial Retrieval}, a higher-order layer that challenges the model to solve multi-step reasoning, complex tasks.

We empirically validate our approach across four topologically distinct urban environments: New York City, Paris, Christchurch, and Manila. These regions were selected to test generalization across diverse road network structures, from the rigid orthogonal grids of the Global North to the irregular, organic layouts often found in the Global South. We train multiple foundation models, including Llama 3.1, Mistral 7B, and Qwen 8B, to assess cross-model robustness. Performance is benchmarked against strong geospatial baselines, like CityGPT \cite{cityGPT:25}, general-purpose foundation models, and inference-time strategies such as Chain-of-Thought (CoT) and few-shot prompting. Our results demonstrate that models trained on our geospatial representations consistently outperform all baselines.

%% file: related_work.tex
\section{Related Work}

Recent studies show that state-of-the-art LLMs retain coarse geographic priors but struggle with basic spatial understanding. 
% Evaluations by \cite{bhandari:23} find that models fail at interpreting coordinates and judging relative positions between urban entities, while \cite{yuhan_ji:25} show inconsistent behavior on topological relations such as containment or intersection. Broader evaluations in \cite{gpt4geo:23} document failures in distance estimation and route reasoning, and embedding analyses from \cite{space_time:24} suggest that latent spatial structures remain too weakly organized for reliable inference. 
% Models often fail to interpret geographic coordinates or accurately determine relative positions between urban entities~\cite{bhandari:23}. Performance is also inconsistent when reasoning about basic topological relationships, such as containment and intersection~\cite{yuhan_ji:25}. Additional evaluations reveal weaknesses in tasks requiring distance estimation and route planning~\cite{gpt4geo:23}. Analyses of model embeddings further suggest that spatial representations in latent space remain insufficiently structured to support reliable spatial inference~\cite{space_time:24}.
Common failures include interpreting geographic coordinates, judging relative positions between urban entities, and reasoning about topological relations such as containment or intersection~\cite{bhandari:23,Ji:25}. Additional evaluations reveal weaknesses in distance estimation and route planning, while embedding analyses suggest that spatial representations in latent space remain insufficiently structured for reliable spatial inference~\cite{gpt4geo:23,space_time:24}. 
These limitations highlight a core challenge: general-purpose models excel at semantic fluency but lack grounded spatial reasoning when queries require combining coordinates, geometry, and constraints.

To address these weaknesses, researchers have begun incorporating structured geospatial data through prompt augmentation \cite{manvi:24}, synthetic instruction tuning \cite{chatmap:23}, and coordinate-aware pretraining \cite{spabert:22, geolm:23}. Domain-specialized models further highlight the importance of structured priors. K2 \cite{k2:24} and ERNIE-GeoL \cite{ernie_geol:22} integrate geoscience graphs, while CityGPT \cite{cityGPT:25} and LAMP \cite{lamp:24} focus on urban datasets and POI-centric tasks. Although these approaches improve city-scale reasoning, they primarily focus on coarse POI metadata or textual summaries. They generally lack the fine-grained spatial operators required to manipulate coordinates, bounding boxes, directions, or adjacency relations.

A separate research direction enhances LLMs through external GIS tools. GeoGPT \cite{geogpt:24} translates natural language into GIS operations, and UrbanLLM \cite{UrbanLLM:24} uses a multi-agent framework to delegate sub-tasks to specialized solvers. While effective when infrastructure is stable, these systems depend on reliable tool access and precise parameterization, which may be unavailable in fast-changing disaster settings. Our approach is complementary. By internalizing core spatial principles and learning structured GIS-aligned representations, we enable models to perform consistent multi-step spatial reasoning both alongside external tools and in scenarios where tool use is constrained.

Finally, specialized non-LLM models like CityFM \cite{cityfm:24} and SARN \cite{sarn:23} provide strong performance on urban mapping and network extraction but do not handle natural-language queries. GeoResponder fills this gap by translating deterministic GIS operations into structured instruction-tuning signals across roads, POIs, and geometric filters. Organized into a three-layer curriculum, these signals strengthen grounding and constraint-aware retrieval. This results in a model that handles a broad suite of disaster-relevant spatial tasks and consistently outperforms both general-purpose LLMs and domain-specific baselines like CityGPT across multiple cities and network topologies.

%% file: methodology.tex
\section{Methodology}
% Note, I will add the new text in blue

% ------------------- cleaned up methodology ---------------------
% We argue that robust geospatial intelligence cannot emerge from unstructured textual pre-training alone; rather, it requires a scaffolded progression from static representation (e.g., real-world knowledge of road networks) to dynamic inference. To this end, we introduce \textbf{GeoResponder}, a framework designed to construct spatial competency from first principles.
% Our approach begins by anchoring semantic entities, such as road segments and critical infrastructure, to the continuous coordinate manifold. This helps the model to establish a grounded mental map of the physical environment. Building upon this foundation, we enforce the internalization of geometric and topological knowledge to the model to derive spatial relationships. Finally, these primitives are synthesized to facilitate high-order reasoning by empowering the model to resolve the multi-step, constraint-heavy optimization problems characteristic of disaster response scenarios. The following sections detail our representation formulations.

\begin{figure}[t]
    \centering
    \includegraphics[width=1\linewidth]{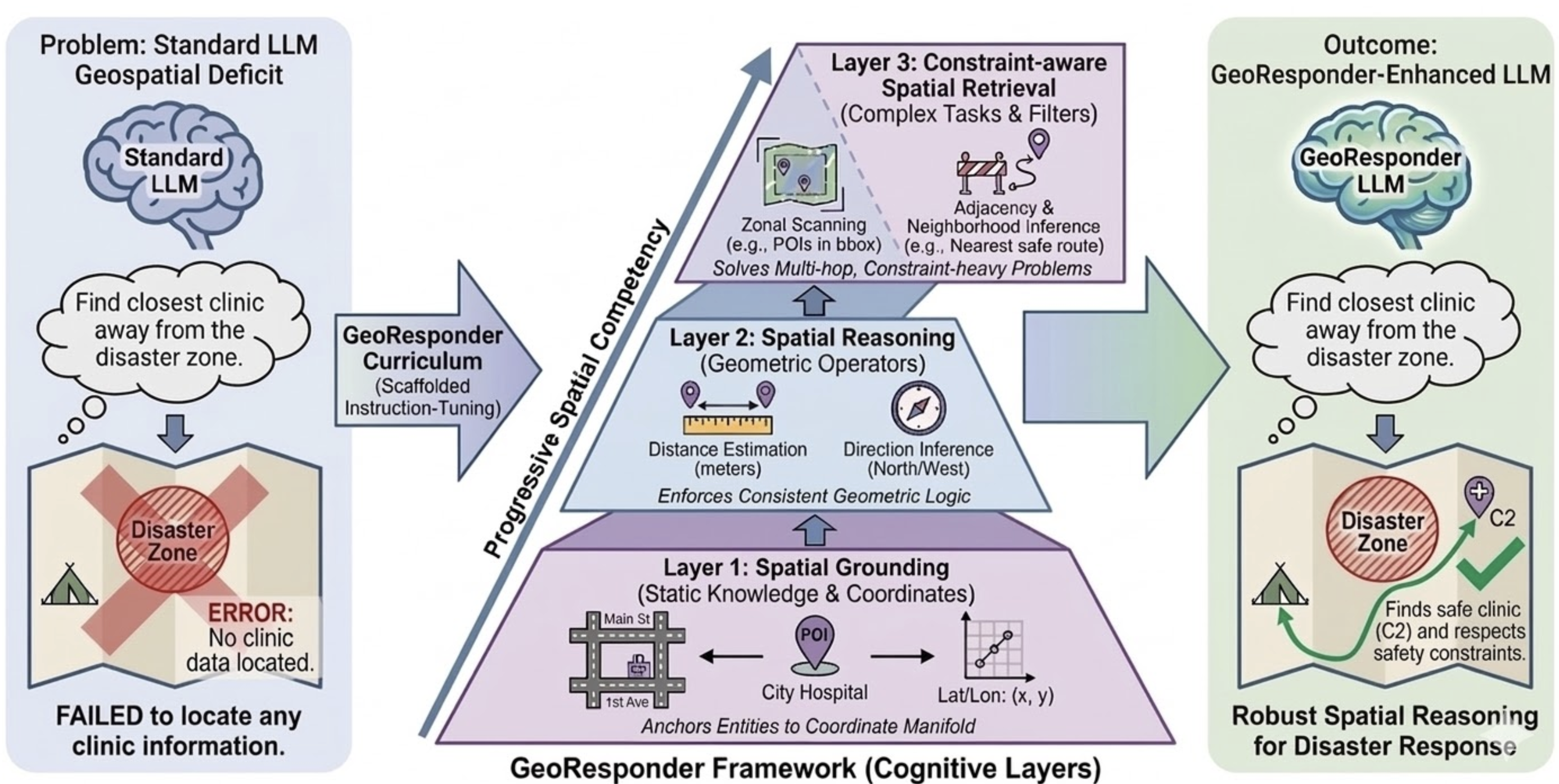}
    \caption{High-level overview of the GeoResponder framework}
    \label{fig:georesponder}
\end{figure}

Robust geospatial intelligence cannot emerge from unstructured textual pre-training alone. Instead, it requires a structured progression from static spatial knowledge to dynamic reasoning. We introduce \textbf{GeoResponder} (Fig. \ref{fig:georesponder}), a framework that constructs spatial competency through a scaffolded curriculum. The approach anchors semantic entities such as road segments and critical infrastructure to the continuous coordinate manifold, allowing the model to build a grounded representation of the physical environment. On top of this foundation, the model learns geometric and topological operators that govern spatial relationships. These primitives are then combined to support multi-step reasoning required to resolve constraint-heavy disaster response queries. The following sections describe the geospatial representations used to train the model.

\input{osm_prelim}
\input{data_representations}

%% file: data_representations.tex
\subsection{Geospatial Representations}
\label{method:representations}

%{\color{blue}
% To enable language models to perform reliable geospatial reasoning and support complex disaster-oriented tasks, we design a set of structured geospatial representations that progressively build three layers of spatial intelligence (Figure \ref{fig:georesponder}). The first layer, \textbf{Spatial Grounding}, teaches the model to internalize a city's geography by linking roads and critical POIs to the continuous coordinate space. This establishes a foundation where the model can interpret, reference, and reason about latitude–longitude pairs and associate them with real locations. The second layer, \textbf{Spatial Reasoning}, introduces fundamental operations over this grounded space, including distance estimation, direction inference, and basic topological checks. Once the model understands where things are and how spatial relationships behave, it must still solve real-world disaster queries that require filtering, constraint handling, and multi-step inference. We therefore introduce a third layer, \textbf{Constraint-aware Spatial Retrieval}, which trains the model to combine grounded knowledge and spatial operations to execute high-level tasks that involve multiple constraints and generalize to unseen scenarios.

% we add notation as a paragraph here
We use OpenStreetMap (OSM) as the source of road-network and point-of-interest (POI) data. For POIs, we focus on critical infrastructure relevant to disaster response (e.g., hospitals, shelters). Roads are represented as polylines composed of lat-lon coordinates. A road segment \(s\) has attributes \(\mathrm{Attr}(s)\) such as road type, speed limit, and length, and is represented by a polyline geometry \(g = [\textit{Coord}_1, \ldots, \textit{Coord}_m]\), where each coordinate \(Coord=(lat,lon)\). A named road $R$ is defined as the set of its constituent segments. POIs are represented by
their coordinates, name, and category $cat$. We also use bounding boxes
$bbox=(lat_{min},lon_{min},lat_{max},lon_{max})$, cardinal directions $dir$ (north, north-east, etc.), and distances
$dist$ measured in meters.

To enable language models to perform reliable geospatial reasoning for disaster-oriented tasks, we design a set of structured geospatial representations that progressively build three layers of spatial intelligence (Figure \ref{fig:georesponder}). The first layer, \textbf{Spatial Grounding}, teaches the model to internalize a city's geography by linking roads and critical POIs to the coordinate space, allowing it to interpret and reference latitude-longitude pairs. The second layer, \textbf{Spatial Reasoning}, introduces fundamental spatial operations such as distance estimation and direction inference. Finally, the third layer, \textbf{Constraint-aware Spatial Retrieval}, trains the model to combine grounded knowledge and spatial operations to solve multi-step complex disaster queries and generalize to unseen scenarios. Next, we elaborate on the representations in detail. 

\subsubsection{Spatial Grounding}
%{\color{blue}
%The first cognitive layer, \emph{spatial grounding}, 
% Through this layer, we provide the foundational factual knowledge and coordinate awareness upon which all higher-level geospatial reasoning is built. The goal of this layer is to make the model internalize the static structure of a city and link it to the continuous coordinate space. This establishes a bidirectional alignment between symbolic descriptions (road names, POI names, categories%\(\mathrm{cat}\)
% , attributes) %\(\mathrm{Attr}(\cdot)\))
% and their geographic representation.

This layer provides the foundational knowledge that links city entities to the coordinate space. The model learns the names and attributes of roads and POIs and how they map to geographic coordinates. This establishes a bidirectional alignment between symbolic descriptions, such as road names or POI categories, and their spatial representation.

To construct this layer, we design a set of atomic representations clustered into two groups. The first group, \emph{Network Topology Encoding}, captures the structure and attributes of the road network. It consists of three atomic representations: \emph{(i) Road Attribute Retrieval} trains the model to map a road name \(R\) to its aggregated attributes \(\mathrm{Attr}(R)\). \emph{(ii) Coordinate Localization} takes a coordinate \textit{Coord} and requires the model to identify the road segment \(s\) for which \textit{Coord} lies on its geometry \(g = [\textit{Coord}_1, \ldots, \textit{Coord}_m]\), returning both \(g\) and \(\mathrm{Attr}(s)\). This enforces a direct grounding between continuous coordinates and the road graph. \emph{(iii) Segment Attribute Inference} uses the geometry \(g\) alone as input and trains the model to infer semantic properties such as road type, speed limit, or lane count, enabling it to understand spatial patterns directly from geometry.

The second Representation group, \emph{POI Coordinate Resolution}, focuses on grounding critical infrastructure entities. \emph{(i) POI Lookup} maps a POI name to its coordinate and category \(cat\), while \emph{(ii) Reverse POI Lookup} maps from a coordinate \textit{Coord} back to the corresponding POI entity. Together, these representations give the model a structured grounding signal: it learns the set of city-specific entities, their attributes \(\mathrm{Attr}(\cdot)\), and how they are embedded in the continuous coordinate space. This layer forms the essential substrate on which subsequent spatial reasoning and complex retrieval tasks are built.
%}

\subsubsection{Spatial Reasoning}
%{\color{blue}
Once the model has acquired spatial grounding and can map between city entities and the coordinate space, the next step is to introduce foundational operations that govern spatial relationships. This layer trains the model to understand how coordinates relate to one another through basic geometric functions. By operating directly on coordinate pairs, the model develops an internal sense of spatial structure beyond memorized facts.

All atomic representations in this layer fall under a single representation group that we refer to as \emph{Geometric Operators}. Within this group, we construct two core atomic representations. \emph{(i) Distance Estimation} presents the model with a pair of coordinates and requires it to predict the geographic distance \(\mathrm{dist}(\textit{Coord}_1, \textit{Coord}_2)\) in meters. This teaches the model how spatial separation behaves in the latitude–longitude system and provides an implicit understanding of how far roads and POIs lie from one another. \emph{(ii) Direction Inference} complements this by training the model to infer the relative position of \(\textit{Coord}_2\) with respect to \(\textit{Coord}_1\), producing a cardinal direction \(\mathrm{dir}\) such as north or west. Together, these representations equip the model with the basic geometric operators required to reason over continuous space and form the basis for more complex constraint-based spatial inference.

\subsubsection{Constraint-aware Spatial Retrieval}
To be useful in disaster settings, a model must solve geospatial tasks that combine grounding, geometric reasoning, filtering, and constraints. The third layer introduces representations that cover these tasks. These representations require the model to perform multi-step inference, integrate multiple forms of spatial information, and generalize beyond the atomic operations learned in earlier layers.

% We cluster the atomic representations in this layer into two groups: \emph{Zonal Scanning} and \emph{Adjacency \& Neighborhood Inference} (Figure ~\ref{fig:georesponder}). \emph{Zonal Scanning} includes three bounding-box based atomic representations as follows: \emph{(i) POI containment:} this representation %requires the model to 
% lists all POIs whose coordinates fall within a given \textit{bbox}, reinforcing the understanding of spatial clustering and inclusion. \emph{(ii) Road containment:} this extends the reasoning to linear features, such as roads that contain at least one segment \(s\) whose geometry \(g = [\textit{Coord}_1,\ldots,\textit{Coord}_m]\) intersects the region defined by the \textit{bbox}. This representation helps the model to reason not only about individual coordinates, but about the spatial extent of a polyline relative to a bounded query region. \emph{(iii) Category Scan:} this representation increases the difficulty by adding a categorical filter, i.e., given a \textit{bbox} and a category \(cat\), the model must retrieve only the POIs of that type inside the region. These atomic representations simulate the kinds of map-window queries used routinely in GIS systems during disaster response (e.g., given the area of a wildfire, list all schools affected by it).

We cluster the atomic representations in this layer into two groups: \emph{Zonal Scanning} and \emph{Adjacency \& Neighborhood Inference} (Figure~\ref{fig:georesponder}). \emph{Zonal Scanning} includes three bounding-box based atomic representations: \emph{(i) POI containment:} this representation lists all POIs whose coordinates fall within a given \textit{bbox}, reinforcing the understanding of spatial clustering and inclusion. \emph{(ii) Road containment:} this extends reasoning to linear features, such as roads that contain at least one segment \(s\) whose geometry \(g = [\textit{Coord}_1,\ldots,\textit{Coord}_m]\) intersects the region defined by the \textit{bbox}. This representation helps the model reason not only about individual coordinates but about the spatial extent of a polyline relative to a bounded query region. \emph{(iii) Category Scan:} this representation increases difficulty by adding a categorical filter, i.e., given a \textit{bbox} and a category \(cat\), the model retrieves only POIs of that type inside the region. These atomic representations simulate map-window queries used routinely in GIS systems during disaster response (e.g., given the area of a wildfire, list all schools affected by it).

% The second group, \emph{Adjacency \& Neighborhood Inference}, covers nearest-entity retrieval and directional neighborhood reasoning. \emph{(i) Directional nearest road}, for a given coordinate \textit{Coord}, contains the closest road segment lying in a specified cardinal direction \(dir\), integrating distance estimation, directional reasoning, and projection onto the road network. \emph{(ii) Nearest POI by category} represents situations when the need is to identify the nearest POI of a certain category. Both atomic representations require multi-hop inference over the grounded spatial knowledge acquired in earlier layers.

The second group, \emph{Adjacency \& Neighborhood Inference}, covers nearest-entity retrieval and directional neighborhood reasoning. \emph{(i) Directional nearest road}, for a given coordinate \textit{Coord}, returns the closest road segment in a specified cardinal direction \(dir\), integrating distance estimation, directional reasoning, and projection onto the road network. \emph{(ii) Nearest POI by category} represents cases where the goal is to identify the nearest POI of a given category. Both representations require multi-hop inference over spatial knowledge acquired in earlier layers.

Together, these representations compel the model to move beyond memorized geospatial facts and simple operations, enabling it to perform structured, constraint-aware retrieval required in realistic disaster-response scenarios.

%% file: datasets.tex
\section{Experimental Setup}

\subsection{Dataset}

% -------------- Figures 1: map figures of the cities
% \begin{figure}[t]
%     \centering
%     \begin{minipage}{0.48\linewidth}
%         \centering
%         \includegraphics[width=\linewidth]{figures/cities/chnz_map.png}
%         \vspace{2pt}
%         {\small Christchurch}
%     \end{minipage}
%     \hfill
%     \begin{minipage}{0.48\linewidth}
%         \centering
%         \includegraphics[width=\linewidth]{figures/cities/manila_map.png}
%         \vspace{2pt}
%         {\small Manila}
%     \end{minipage}

%     \vspace{6pt}

%     \begin{minipage}{0.48\linewidth}
%         \centering
%         \includegraphics[width=\linewidth]{figures/cities/paris_map.png}
%         \vspace{2pt}
%         {\small Paris}
%     \end{minipage}
%     \hfill
%     \begin{minipage}{0.48\linewidth}
%         \centering
%         \includegraphics[width=\linewidth]{figures/cities/nyc_map.png}
%         \vspace{2pt}
%         {\small New York City}
%     \end{minipage}

%     \caption{Maps of the four benchmark cities with grey roads and colored POI dots, where each color indicates a different POI category.}

%     \label{fig:city_maps}
% \end{figure}

To assess generalization, we select four cities: Christchurch (New Zealand), Manila (Philippines), Paris (France), and New York City (United States) from four continents. Christchurch and Manila are prone to disasters such as earthquakes and typhoons.
Paris and New York City represent dense and globally recognizable urban environments with distinct spatial layouts and linguistic contexts.
Table~\ref{tab:city_stats_training} highlights the main attributes of each city.
Paris has the highest concentration of POIs, while Christchurch has the lowest. Manila and New York City exhibit comparably dense and heterogeneous urban environments. Paris contains 4,062 critical POIs (about 38 per km\(^2\)), while Christchurch has 606. Manila and New York City show similar POI densities relative to their larger areas. Across all cities, the distribution of critical-infrastructure categories is similar. 

% ------------- Table 2: city osm stats
% \begin{table}[!t]
% \centering
% \caption{City-level OpenStreetMap statistics.}
% \label{tab:city_stats}
% \small
% \setlength{\tabcolsep}{3pt}

% \begin{tabular}{lcccc}
% \toprule
% \textbf{Statistic} &
% \textbf{Christch.} &
% \textbf{Paris} &
% \textbf{Manila} &
% \textbf{New York} \\
% \midrule
% Continent & Oceania & Europe & Asia & N.\ America \\
% City area (km$^2$) & 295 & 105 & 619 & 778 \\
% Road segments & 23{,}773 & 17{,}681 & 124{,}275 & 135{,}625 \\
% Unique roads (name) & 3{,}980 & 3{,}865 & 11{,}514 & 7{,}776 \\
% Total road length (km) & 4{,}868 & 2{,}634 & 12{,}485 & 20{,}619 \\
% Road length density (km/km$^2$) & 16.5 & 25.1 & 20.2 & 26.5 \\
% \# of POIs (Critical Infrastructure) & 606 & 4{,}062 & 7{,}015 & 5{,}586 \\
% POI density (per km$^2$) & 2.1 & 38.7 & 11.3 & 7.2 \\
% \bottomrule
% \end{tabular}
% \end{table}

\begin{table}[t]
\centering
\caption{City statistics and training data sizes used in GeoResponder.}
\label{tab:city_stats_training}

\footnotesize
\setlength{\tabcolsep}{3pt}
\renewcommand{\arraystretch}{0.95}

\begin{tabular}{lrrrr}
\toprule
 & \textbf{Christch.} & \textbf{Paris} & \textbf{Manila} & \textbf{New York} \\
\midrule

\multicolumn{5}{l}{\textbf{City characteristics}} \\
Continent & Oceania & Europe & Asia & N.\ America \\
Area (km$^2$) & 295 & 105 & 619 & 778 \\
Road segments & 23{,}773 & 17{,}681 & 124{,}275 & 135{,}625 \\
Unique roads & 3{,}980 & 3{,}865 & 11{,}514 & 7{,}776 \\
Road length (km) & 4{,}868 & 2{,}634 & 12{,}485 & 20{,}619 \\
Road density & 16.5 & 25.1 & 20.2 & 26.5 \\
POIs (critical) & 606 & 4{,}062 & 7{,}015 & 5{,}586 \\
POI density & 2.1 & 38.7 & 11.3 & 7.2 \\

\midrule

\multicolumn{5}{l}{\textbf{Training data}} \\
Spatial grounding & 100{,}919 & 87{,}517 & 529{,}589 & 567{,}402 \\
Spatial reasoning & 4{,}800 & 4{,}800 & 4{,}800 & 4{,}800 \\
Constraint retrieval & 28{,}566 & 28{,}010 & 32{,}045 & 48{,}530 \\

\midrule
\textbf{Total samples} & 134{,}285 & 120{,}327 & 566{,}434 & 620{,}732 \\
\bottomrule
\end{tabular}

\end{table}

%\FloatBarrier

\subsubsection{Training Data}
%{\color{blue}

For each city, we generate training data by instantiating the three cognitive layers of geospatial representations described in Section~\ref{method:representations}. All instances from these layers are then combined into a single city-specific dataset. In addition to the standard instruction-style formats, we also include a subset of multiple-choice (MCQ) variants to increase prompt diversity and better align with MCQ-style evaluation tasks. Table~\ref{tab:city_stats_training} reports important stats of training data. Christchurch and Paris contain approximately 120--135k training instances, whereas Manila and New York City range between 500k and 600k. The majority of this increase arises from the spatial grounding layer, which scales with city area and POI density, producing substantially larger grounding datasets for Manila and New York compared to Paris and Christchurch.
%}

%\subsubsection{Sampling Strategy}
%{\color{blue}
Several representations require sampled coordinates (e.g., directional nearest–road, nearest POI by category), while others require sampled bounding boxes (e.g, POI containment). For coordinates, we use a density-aware strategy: the area of interest (the city) is divided into a uniform grid, and the number of points drawn per cell is proportional to the local density of road segments.
Bounding boxes are sampled differently. We draw boxes with a maximum allowed area and enforce a minimum aspect-ratio constraint to avoid degenerate, overly thin regions. To ensure coverage of empty zones, we also include a controlled number of \emph{empty} boxes that contain no POIs or road segments, which helps the model learn to distinguish between populated and non-populated spatial windows.
%}

% \subsection{Evaluation Tasks}

% ----------------- task definition table
\begin{table*}[t]
\centering
\scriptsize
\setlength{\tabcolsep}{4pt}
\renewcommand{\arraystretch}{1.03}

\caption{Taxonomy of evaluation tasks across the three cognitive layers.}
\label{tab:eval_tasks}

\begin{tabular}{
    m{2.35cm}
    >{\centering\arraybackslash}m{2.75cm}
    m{4.6cm}
    >{\centering\arraybackslash}m{1.05cm}
}
\toprule
\textbf{Task Name} &
\textbf{Notation} &
\textbf{Disaster Response Example} &
\textbf{Metric} \\
\midrule

\multicolumn{4}{l}{\textbf{\textcolor{colL1}{1. Spatial Grounding}} {\scriptsize\textit{\textcolor{colL1}{(Static Knowledge)}}}} \\
\midrule
\textcolor{colL1}{Rev. Road Attr. Lookup}
& $Coord (\in Seg) \rightarrow Attr$
& Check if a road segment supports heavy rescue vehicles.
& Acc, MAPE, F1 \\

\textcolor{colL1}{Rev. Road Lookup}
& $Coord (\in Seg) \rightarrow Road$
& Retrieve street name to guide ground teams.
& Acc \\

\textcolor{colL1}{POI Lookup}
& $POI \rightarrow Coord$
& Convert “Fire at City Hospital” into coordinates.
& F1 \\

\textcolor{colL1}{Rev. POI Lookup}
& $Coord \rightarrow POI$
& Identify a collapsed facility from aerial coordinates.
& Acc \\

\midrule
\multicolumn{4}{l}{\textbf{\textcolor{colL2}{2. Spatial Reasoning}} {\scriptsize\textit{\textcolor{colL2}{(Geometric Logic)}}}} \\
\midrule
\textcolor{colL2}{Distance Est.}
& $(Coord_A, Coord_B) \rightarrow Meters$
& Determine if a supply truck can reach a shelter with fuel.
& MAPE \\

\textcolor{colL2}{Direction Inf.}
& $(Coord_A, Coord_B) \rightarrow Dir.$
& Determine a safe helicopter heading to avoid smoke.
& F1 \\

\midrule
\multicolumn{4}{l}{\textbf{\textcolor{colL3}{3. Constraint-Aware Spatial Retrieval}} {\scriptsize\textit{\textcolor{colL3}{(Constraint Inf.)}}}} \\
\midrule
\textcolor{colL3}{Road Containment}
& $BBox \rightarrow \{Roads_{in}\}$
& List roads within the predicted flood polygon.
& F1 \\

\textcolor{colL3}{Road Exclusion}
& $BBox \rightarrow \{Roads_{out}\}$
& Identify staging areas outside a chemical spill.
& -- \\

\textcolor{colL3}{POI Containment}
& $BBox \rightarrow \{POIs_{in}\}$
& List schools located within the earthquake area.
& F1 \\

\textcolor{colL3}{POI Exclusion}
& $BBox \rightarrow \{POIs_{out}\}$
& Identify shelters outside the disaster area.
& -- \\

\textcolor{colL3}{Category Scan}
& $(BBox, Cat) \rightarrow \{POI_{list}\}$
& Retrieve all gas stations in a grid sector to secure fuel supply.
& F1 \\

\textcolor{colL3}{Nearest POI by Cat}
& $(Coord, Cat) \rightarrow POI$
& Locate the nearest pharmacy to a location.
& Acc \\

\textcolor{colL3}{Neighbor POI}
& $POI_A \rightarrow POI_B$
& Find the closest hospital to the school.
& -- \\

\textcolor{colL3}{Nearest Road (POI)}
& $POI \rightarrow Road_{nearest}$
& Find the nearest road access point near an evacuation camp.
& -- \\

\textcolor{colL3}{Dir. Nearest Road}
& $(Coord, Dir) \rightarrow Road$
& Find the nearest safe road north of some coordinates.
& Hit@K, Acc@1km, MRR \\

\textcolor{colL3}{Nearest Road}
& $Coord \rightarrow Road$
& Identify the nearest road to some coordinates.
& Hit@K, Acc@1km, MRR \\

\bottomrule
\end{tabular}
\end{table*}

\subsection{Evaluation Tasks}
% To assess geospatial reasoning performance, we evaluate our models on a suite of downstream tasks that are both challenging and directly relevant to disaster response. Table~\ref{tab:eval_tasks} lists all evaluation tasks, organized by the cognitive layer they probe and by the underlying geospatial operation they correspond to. For clarity, the table also specifies the input--output notation for each task and provides a short disaster-response scenario illustrating its real-world relevance.

% We report results for both multiple-choice (MCQ) and free-form versions of tasks. The MCQ format supplies four candidate answers, while the free-form setting requires the model to generate the correct output without any options provided, making it substantially more difficult.

% To further test the limits of generalization, we introduce a set of out-of-distribution (OOD) tasks representing geospatial operations that are never seen during training. These include \emph{POI Exclusion}, \emph{Road Exclusion}, \emph{Neighbor POI Identification}, and \emph{POI-to-Nearest-Road}. When evaluated, each OOD task is framed within a realistic disaster-response scenario. This setup allows us to measure whether the model can transfer its learned spatial reasoning capabilities to entirely novel geospatial queries.

To assess geospatial reasoning, we evaluate models on a suite of downstream tasks that are both challenging and relevant to disaster response. Table~\ref{tab:eval_tasks} lists all tasks, organized by cognitive layer and underlying geospatial operation. For clarity, the table also specifies the input--output notation and provides a short disaster-response scenario illustrating each task’s real-world relevance.

We report results for both MCQs and free-form versions. The MCQ format provides four candidate answers, while the free-form setting requires the model to generate the correct output without options, making it more challenging. To further test generalization, we include several out-of-distribution (OOD) tasks representing geospatial operations \emph{not seen} during training: \emph{POI Exclusion}, \emph{Road Exclusion}, \emph{Neighbor POI Identification}, and \emph{POI-to-Nearest-Road}. Each OOD task is framed within a realistic disaster-response scenario.

\subsection{Evaluation Metrics}

% We employ a diverse set of evaluation metrics strictly aligned with the output modality of each geospatial task. While standard accuracy suffices for multiple-choice questions, we evaluate free-form geometric and retrieval tasks using specialized rank-aware and distance-based measures. For distance estimation, we report Mean Absolute Percentage Error (MAPE), whereas retrieval tasks (e.g., \emph{Nearest Road}) are assessed via \emph{Acc@1km}—verifying if predictions lie within a 1\,km geodesic radius—and \emph{Hit@k}, formally defined as $\text{Hit@}k = \frac{1}{n} \sum_{i=1}^{n} 1\{\hat{r}_i \in G_i^{(k)}\}$, alongside Mean Reciprocal Rank (MRR). For directional variants, these metrics are averaged across all cardinalities. Finally, set-valued generation tasks require overlap-based scoring: \emph{Road Containment} utilizes exact-intersection F1 scores, while \emph{POI Lookup} adopts a greedy one-to-one geodesic matching protocol (Algorithm~\ref{alg:poi_lookup_f1}) that validates predictions falling within a 150\,m threshold of ground-truth coordinates.

We employ a diverse set of evaluation metrics aligned with the output modality of each geospatial task. While standard accuracy suffices for MCQs, we evaluate free-form geometric and retrieval tasks using specialized rank-aware and distance-based measures. For distance estimation, we report Mean Absolute Percentage Error (MAPE), whereas retrieval tasks (e.g., \emph{Nearest Road}) are assessed via \emph{Acc@1km}—verifying if predictions lie within a 1\,km geodesic radius—and \emph{Hit@k}, formally defined as $\text{Hit@}k = \frac{1}{n} \sum_{i=1}^{n} 1\{\hat{r}_i \in G_i^{(k)}\}$, alongside Mean Reciprocal Rank (MRR). For directional variants, these metrics are averaged across cardinalities. Finally, set-valued generation tasks require overlap-based scoring: \emph{Road Containment} utilizes exact-intersection F1 scores, while \emph{POI Lookup} adopts a greedy one-to-one geodesic matching protocol (Algorithm~\ref{alg:poi_lookup_f1}) that validates predictions within a 150\,m threshold of ground-truth coordinates.

% For all MCQ representations we report \textbf{Accuracy}. Free-form tasks use metrics aligned with their output type.

% \paragraph{Distance and Ranking Metrics.}
% For distance estimation we use MAPE. Nearest-road and directional-nearest-road tasks additionally use \emph{Acc@1km}, which counts a prediction as correct if the predicted road lies within 1\,km of the ground-truth location.

% These tasks also use standard ranking metrics. \emph{Hit@k} measures whether the predicted road $\hat{r}_i$ lies among the ground-truth top-$k$ nearest roads:
% \[
% \text{Hit@}k = \frac{1}{n} \sum_{i=1}^{n} 1\{\hat{r}_i \in G_i^{(k)}\}.
% \]
% \emph{MRR} (Mean Reciprocal Rank) assigns higher credit when the predicted road appears earlier in the ground-truth nearest-road ordering. For \emph{Directional Nearest Road}, all metrics are computed for each cardinal direction and averaged. Set-valued representations such as \emph{Road Containment} are evaluated using F1 based on exact set intersection.

% \paragraph{POI Lookup.}
% Because a POI may correspond to multiple valid coordinates and predictions lie in continuous space, we evaluate using greedy one-to-one geodesic matching: a predicted coordinate is correct if it falls within 150\,m of a unique ground-truth coordinate. Precision, recall, and F1 are computed over these matches (Algorithm~\ref{alg:poi_lookup_f1}).

\begin{algorithm}[t]
\caption{F1 Evaluation for POI Lookup}
\label{alg:poi_lookup_f1}
\begin{algorithmic}[1]
\scriptsize
\Require Predicted coordinate sets $\{\hat{C}_i\}$, ground-truth sets $\{C_i\}$, distance threshold $\tau$
\For{$i = 1$ to $N$}
    \State $G \gets$ set of ground-truth coordinates in $C_i$
    \State $n_{\text{correct}} \gets 0$
    \For{each predicted coordinate $\hat{c} \in \hat{C}_i$}
        \State $(g^\star, d^\star) \gets \arg\min_{g \in G} \textsc{GeodesicDistance}(\hat{c}, g)$
        \If{$d^\star \le \tau$}
            \State $n_{\text{correct}} \gets n_{\text{correct}} + 1$
            \State $G \gets G \setminus \{g^\star\}$ \Comment{greedy one-to-one match}
        \EndIf
    \EndFor
    \State $p_i \gets n_{\text{correct}} / |\hat{C}_i|$
    \State $r_i \gets n_{\text{correct}} / |C_i|$
    \State $f_i \gets \frac{2 p_i r_i}{p_i + r_i}$ \Comment{F1 score}
    \State Record $f_i$
\EndFor
\State \Return mean F1 across all examples
\end{algorithmic}
\end{algorithm}
%}

% ---- Baselines
% scratch
% To evaluate Georeaponder, we compare against strong open source instruction tuned models: qwen 3 8b, llama 3.1 8b, and mistral 7b v0.3 as well as GPT-4o Mini. 

% we use two prompting settings, one is direct prompting (simple task specific, without enhancements). another is geo prompt fusion, which is a combination of three main advanced strategies: chain of thought to increase the reaoning capabilties of the models, contextual grounding (injecting structured osm information to give the model more context, as in manvi:24. and QSF learning AEF:25 which retrieves few shot examples that are geographically close to the task, even further helping the model. 

% For the MCQ tasks, we use a state of the art geospatial llm, citygpt cite them here, which we train the qwen 2.5 B version, using the hyperparameters shared in their paper. we train it only on Paris and Nyc as they dont have the dataste on chnz or manila, and we train the sft version of it, due to gpu constraint. their swft method, as they report in the paper, is mainly to generalize to non geospatial domains, and their sft and swft results are near identical on geosptial tasks, so we stick with the sft version. the swft version is expensive. we dont evalute citygpt on non mcq results as we found it is not able to answer these tasks (near 0 on all non mcq metrics)

% ------------- Table 2: Mcq In distribution questions
\begin{table}[htb]
\centering
\caption{Performance across six geospatial tasks for GeoResponder and CityGPT. Best GeoResponder values (Mistral) are bold.}
\label{tab:mcq_in_distribution}
\renewcommand{\arraystretch}{1.15}
\resizebox{\textwidth}{!}{
\begin{tabular}{ll|c|c|c|c|c|c}
\toprule
City & Model &
\multicolumn{1}{c|}{POI Lookup} &
\multicolumn{1}{c|}{Rev. POI Lookup} &
\multicolumn{1}{c|}{Road Containment} &
\multicolumn{1}{c|}{POI Containment} &
\multicolumn{1}{c|}{Nearest POI by Cat} &
\multicolumn{1}{c}{Category Scan} \\
\midrule

% ------------------ Christchurch ------------------
\multirow{3}{*}{Christchurch}
& GeoResponder (LLaMa)   & 0.5   & 0.564 & 0.574 & 0.694 & 0.7435 & 0.689 \\
& GeoResponder (Mistral) & \textbf{\underline{0.78}}  & \textbf{\underline{0.637}} & \textbf{\underline{0.637}} & \textbf{\underline{0.793}} & \textbf{\underline{0.856}}  & \textbf{\underline{0.77}}  \\
& GeoResponder (Qwen)    & 0.472 & 0.317 & 0.536 & 0.565 & 0.4    & 0.561 \\
\midrule

% ------------------ Paris ------------------
\multirow{6}{*}{Paris}
& CityGPT         & 0.316 & 0.219 & 0.291 & 0.223 & 0.24  & 0.259 \\
\cmidrule(lr){2-8}
& GeoResponder (LLaMa)   & 0.624 & 0.425 & 0.601 & 0.583 & 0.467 & 0.505 \\
& GeoResponder (Mistral) & \textbf{\underline{0.75}}  & \textbf{\underline{0.62}}  & \textbf{\underline{0.68}}  & \textbf{\underline{0.685}} & \textbf{\underline{0.723}} & \textbf{\underline{0.633}} \\
& GeoResponder (Qwen)    & 0.46  & 0.25  & 0.456 & 0.482 & 0.326 & 0.345 \\
\cmidrule(lr){2-8}
& \textbf{Best GeoResponder} 
& 0.75  & 0.62  & 0.68  & 0.685 & 0.723 & 0.633 \\
& \textbf{vs CityGPT (\%)} 
& \multicolumn{1}{c|}{\cellcolor{green!40}{+137}} 
& \multicolumn{1}{c|}{\cellcolor{green!40}{+183}} 
& \multicolumn{1}{c|}{\cellcolor{green!40}{+134}} 
& \multicolumn{1}{c|}{\cellcolor{green!40}{+207}} 
& \multicolumn{1}{c|}{\cellcolor{green!40}{+201}} 
& \multicolumn{1}{c}{\cellcolor{green!40}{+144}} \\
\midrule

% ------------------ Manila ------------------
\multirow{3}{*}{Manila}
& GeoResponder (LLaMa)   & 0.488 & 0.43  & 0.512 & 0.509 & 0.47  & 0.564 \\
& GeoResponder (Mistral) & \textbf{\underline{0.497}} & \textbf{\underline{0.5}}   & \textbf{\underline{0.576}} & \textbf{\underline{0.576}} & \textbf{\underline{0.562}} & \textbf{\underline{0.595}} \\
& GeoResponder (Qwen)    & 0.464 & 0.419 & 0.5   & 0.508 & 0.432 & 0.503 \\
\midrule

% ------------------ New York City ------------------
\multirow{6}{*}{New York City}
& CityGPT         & 0.33  & 0.247 & 0.31  & 0.23  & 0.265 & 0.274 \\
\cmidrule(lr){2-8}
& GeoResponder (LLaMa)   & 0.46  & 0.45  & 0.6   & 0.61  & 0.6   & 0.61  \\
& GeoResponder (Mistral) & \textbf{\underline{0.517}} & \textbf{\underline{0.568}} & \textbf{\underline{0.685}} & \textbf{\underline{0.714}} & \textbf{\underline{0.749}} & \textbf{\underline{0.723}} \\
& GeoResponder (Qwen)    & 0.437 & 0.394 & 0.602 & 0.632 & 0.476 & 0.614 \\
\cmidrule(lr){2-8}
& \textbf{Best GeoResponder} 
& 0.517 & 0.568 & 0.685 & 0.714 & 0.749 & 0.723 \\
& \textbf{vs CityGPT (\%)} 
& \multicolumn{1}{c|}{\cellcolor{green!40}{+57}} 
& \multicolumn{1}{c|}{\cellcolor{green!40}{+130}} 
& \multicolumn{1}{c|}{\cellcolor{green!40}{+121}} 
& \multicolumn{1}{c|}{\cellcolor{green!40}{+210}} 
& \multicolumn{1}{c|}{\cellcolor{green!40}{+183}} 
& \multicolumn{1}{c}{\cellcolor{green!40}{+164}} \\
\bottomrule
\end{tabular}
}
\end{table}

%{\color{blue}
\subsection{Baselines}

We compare GeoResponder against strong instruction-tuned baselines such as Qwen~3~8B, LLaMA~3.1~8B, Mistral~7B~v0.3, GPT-4o~Mini. All models are evaluated under two prompting regimes: (i) \emph{Direct Prompting}, using concise task-specific instructions; and (ii) \emph{Geo Prompt Fusion}, which integrates chain-of-thought reasoning, structured OSM-based contextual grounding (similar to ~\cite{manvi:24}), and QSF-based retrieval of geographically proximate few-shot exemplars~\cite{AEF:25}.

For \emph{multiple-choice} tasks, we also benchmark against the geospatial LLM \textbf{CityGPT}~\cite{cityGPT:25} using their SFT variant. We reproduce their Qwen~2.5~7B model using the hyperparameters reported in their paper. Although CityGPT provides training data for several cities, it does not release datasets for two of our evaluation cities (Christchurch and Manila), so we train it only on Paris and NYC. %We use their SFT variant (not SWFT) due to GPU constraints; as reported in their work, SFT and SWFT achieve nearly identical performance on geospatial tasks, while SWFT incurs substantially higher cost.

%% file: evaluation_and_results.tex
\section{Evaluation and Results}

% ----------- Figure 6: Disaster OOD GeoResponder vs CityGPT
% \begin{figure}[htbp]
%     \centering
%     \includegraphics[width=0.6\columnwidth]{figures/disaster_ood_tasks.pdf}

%     \caption{
%         Out-of-distribution disaster tasks: comparison between CityGPT performance 
%         (grey) and the improvement achieved by GeoResponder–Mistral (yellow). 
%         %Each city panel shows four task categories. GeoResponder consistently outperforms CityGPT across all tasks in both Paris and New York City.
%     }

%     \label{fig:disaster_ood_tasks}
% \end{figure}

\subsection{In-Distribution MCQ Tasks}
%{\color{blue}
Table~\ref{tab:mcq_in_distribution} reports results on the in-distribution MCQ tasks. In Paris and New York, we additionally compare GeoResponder against CityGPT.
CityGPT performs well on simpler grounding tasks such as POI Lookup and Road Containment, but its accuracy drops sharply on tasks that depend on coordinates or reasoning over bounded regions (Reverse POI Lookup, POI Containment, Nearest-POI-by-Category, Category Scan). This highlights its limited understanding of fine-grained spatial relationships despite being pre-trained on OSM-derived data.

GeoResponder, particularly the Mistral variant, achieves large gains across all tasks and all cities. Improvements over CityGPT typically exceed \textbf{+100\%} and often surpass \textbf{+150\%}, with the largest margins appearing precisely on the tasks where CityGPT struggles. This demonstrates GeoResponder’s stronger grounding in spatial neighborhoods, containment logic, and category-filtered retrieval.

Absolute accuracy varies by city. Manila and New York, which cover larger areas and exhibit high POI diversity, yield lower scores on POI Lookup and Reverse POI Lookup due to increased spatial ambiguity and denser candidate sets. Christchurch and Paris show higher accuracy on grounding-oriented tasks, consistent with their smaller spatial extents and more concentrated POI distributions. Overall, these results confirm that (i) structured geospatial representations are essential for reliable spatial reasoning, (ii) GeoResponder generalizes across cities with diverse spatial scales, and (iii) task difficulty correlates strongly with city size, POI density, and the geometric complexity of the underlying operation.

% \begin{figure}[h]
%     \centering
%     \includegraphics[width=0.6\linewidth]{figures/in_distribution_non_mcq_bars.pdf}
%     \caption{In-distribution non-MCQ bar distribution.}
%     \label{fig:in_distribution_non_mcq_bars}
% \end{figure}

\subsection{Road-network reasoning on free-form tasks}
%{\color{blue}
Table~\ref{tab:all_tasks_full} evaluates whether GeoResponder's three-layer curriculum yields reliable free-form geospatial reasoning. Results are reported across four cities and three task groups aligned with our cognitive layers: (i) \emph{Nearest Road Retrieval}, (ii) \emph{Geometric Operators} (distance, direction), and (iii) \emph{Reverse Road Attribute and Name Lookup}. For each city, we compare base instruction-tuned models, their Geo Prompt Fusion variants, and all GeoResponder backbones.

% ------ table 1: non mcq old tasks on four cities. no citygpt here
% important note: in paris, Mistral 7B v0.3 does better in two small tasks, a hypothesis is bias (Mistral 7B v0.3 is a french model)

\begin{table}[t]
\centering
\caption{Performance evaluation across eight tasks and four cities. Darker shades indicate superior performance. %A detailed performance evaluation of all models across the full range of tasks and cities. For each city, the final two rows summarize the results of the best-performing GeoResponder model and compare it to the strongest baseline model. Cells are color-coded by task category, with darker shades indicating superior performance on the specific metric.
}
\resizebox{\textwidth}{!}{
\begin{tabular}{ll| H H H| H H H| K| I| J| M J| J| J}
\toprule
\multirow{3}{*}{City} & \multirow{3}{*}{Model} & 
\multicolumn{6}{c|}{\textbf{Nearest Road}} & 
\multicolumn{2}{c|}{\textbf{Geometric Operators}} & 
\multicolumn{5}{c}{\textbf{Rev. Road Attr. and Name Lookup}} \\
\cmidrule(lr){3-8} \cmidrule(lr){9-10} \cmidrule(lr){11-15}
& & 
\multicolumn{3}{c|}{Standard} & 
\multicolumn{3}{c|}{Directional} & 
\multicolumn{1}{c}{Dist} & \multicolumn{1}{c|}{Dir} & 
\multicolumn{1}{c}{Road} & \multicolumn{2}{c}{Length} & 
\multicolumn{1}{c}{Speed} & 
\multicolumn{1}{c|}{Lanes} \\
\cmidrule(lr){3-5} \cmidrule(lr){6-8} \cmidrule(lr){9-9} \cmidrule(lr){10-10}
\cmidrule(lr){11-11} \cmidrule(lr){12-13} \cmidrule(lr){14-14} \cmidrule(lr){15-15}
& & 
\multicolumn{1}{c}{Hit@1} &
\multicolumn{1}{c}{Acc@1km} &
\multicolumn{1}{c}{MRR} &
\multicolumn{1}{c}{Hit@1} &
\multicolumn{1}{c}{Acc@1km} &
\multicolumn{1}{c}{MRR} &
\multicolumn{1}{c}{MAPE ↓} &
\multicolumn{1}{c|}{F1} &
\multicolumn{1}{c}{Acc } &
\multicolumn{1}{c}{MAPE ↓} &
\multicolumn{1}{c}{Acc@30\%} &
\multicolumn{1}{c}{F1} &
\multicolumn{1}{c|}{F1} \\
\midrule

% ------------------ Christchurch ------------------
\multirow{13}{*}{Christchurch}
& LLaMa 3.1 8B & 0 & 0.013 & 0.002 & 0 & 0 & 0 & 0.695 & 0.14 & 0.002 & 2.7821 & 0.114 & 0 & 0.2332 \\
& Mistral 7B v0.3 & 0.002 & 0.017 & 0.001 & 0.027 & 0.027 & 0.018 & 1.33 & 0.03 & 0 & 8.3 & 0.016 & 0 & 0.1858 \\
& Qwen3 8B & 0 & 0.004 & 0.001 & 0.002 & 0.002 & 0.002 & 0.678 & 0.03 & 0.001 & 1.6582 & 0.235 & 0 & 0.2334 \\
\cmidrule{2-15}

& GeoPromptFusion (LLaMa) & 0.16 & 0.44 & 0.218 & 0.134 & 0.237 & 0.165 & 0.4429 & 0.19 & 0.064 & 0.9429 & 0.289 & 0.1772 & 0.1864 \\
& GeoPromptFusion (Mistral) & 0.169 & 0.595 & 0.252 & 0.092 & 0.124 & 0.101 & 0.6697 & 0.06 & 0.025 & 1.0249 & 0.2683 & 0.1869 & 0.2829 \\
& GeoPromptFusion (Qwen) & 0.23 & 0.676 & 0.321 & 0.129 & 0.265 & 0.172 & 0.37248 & 0.31 & 0.163 & 0.7309 & 0.3023 & 0.2528 & 0.2367 \\
& GeoPromptFusion (GPT-4o mini) & 0.103 & 0.289 & 0.138 & 0.103 & 0.164 & 0.122 & 0.4409 & 0.35 & 0.039 & 1.1068 & 0.286 & 0.1389 & 0.2644 \\
\cmidrule{2-15}
& GeoResponder (LLaMa) & 0.17 & 0.669 & 0.275 & 0.224 & 0.372 & 0.28 & 0.078 & 0.73 & 0.529 & 0.7419 & 0.2735 & 0.521 & 0.2761 \\
& GeoResponder (Mistral) & 0.345 & 0.794 & 0.493 & 0.288 & 0.417 & 0.34 & 0.0615 & 0.94 & 0.74 & 0.5656 & 0.53 & 0.631 & 0.4307 \\
& GeoResponder (Qwen) & 0.026 & 0.21 & 0.056 & 0.03 & 0.089 & 0.047 & 0.167 & 0.9 & 0.047 & 0.769 & 0.14 & 0.16 & 0.179 \\
\cmidrule{2-15}

& \textbf{Best GeoResponder} 
& 0.345 & 0.794 & 0.493 
& 0.288 & 0.417 & 0.34 
& 0.0615 & 0.94 
& 0.74 
& 0.5656 & 0.53 
& 0.631 
& 0.4307 \\
& \textbf{vs Best Baseline (\%)} 
& \multicolumn{1}{c|}{\cellcolor{green!40}{+50}}   % Hit@1 std
& \multicolumn{1}{c|}{\cellcolor{green!40}{+17}}   % Acc@1km std
& \multicolumn{1}{c|}{\cellcolor{green!40}{+54}}   % MRR std
& \multicolumn{1}{c|}{\cellcolor{green!40}{+115}}  % Hit@1 dir
& \multicolumn{1}{c|}{\cellcolor{green!40}{+57}}   % Acc@1km dir
& \multicolumn{1}{c|}{\cellcolor{green!40}{+98}}   % MRR dir
& \multicolumn{1}{c|}{\cellcolor{green!40}{-83}}   % Dist MAPE (↓, worse)
& \multicolumn{1}{c|}{\cellcolor{green!40}{+169}}  % Dir F1
& \multicolumn{1}{c|}{\cellcolor{green!40}{+354}}  % Road Acc
& \multicolumn{1}{c|}{\cellcolor{green!40}{-23}}   % Length MAPE (↓, better)
& \multicolumn{1}{c|}{\cellcolor{green!40}{+75}}   % Length A@30%
& \multicolumn{1}{c|}{\cellcolor{green!40}{+150}}  % Speed F1
& \multicolumn{1}{c}{\cellcolor{green!40}{+52}}    % Lanes F1
\\
\midrule

% ------------------ Paris ------------------
\multirow{13}{*}{Paris}
& LLaMa 3.1 8B   & 0 & 0.005 & 0 & 0 & 0.007 & 0.002 & 0.829 & 0.14 & 0 & 0.9 & 0.3719 & 0.246 & 0.07 \\
& Mistral 7B v0.3 & 0 & 0.003 & 0 & 0 & 0.007 & 0.001 & 0.953 & 0.05 & 0.004 & 20 & 0 & 0.2289 & 0.06 \\
& Qwen3 8B    & 0 & 0.032 & 0.001 & 0.002 & 0.011 & 0.003 & 0.761 & 0.43 & 0.002 & 4.99 & 0.152 & 0.0516 & 0.1 \\
\cmidrule{2-15}

& GeoPromptFusion (LLaMa)   & 0.022 & 0.213 & 0.04 & 0.073 & 0.168 & 0.098 & 0.76 & 0.21 & 0.02 & 0.736 & 0.3375 & 0.257 & 0.3 \\
& GeoPromptFusion (Mistral) & 0.03  & 0.336 & 0.06 & 0.03  & 0.07  & 0.039 & 0.99 & 0.1  & 0.02 & 0.67  & 0.354  & 0.2789 & 0.09 \\
& GeoPromptFusion (Qweni)   & 0.09  & 0.558 & 0.16 & 0.08  & 0.236 & 0.121 & 0.82 & 0.43 & 0.14 & 0.73  & 0.256  & 0.309 & 0.198 \\
& GeoPromptFusion (GPT-4o mini) & 0.032 & 0.281 & 0.06 & 0.022 & 0.093 & 0.04 & 0.48 & 0.33 & 0.03 & 1.06 & 0.315 & 0.0954 & 0.1 \\
\cmidrule{2-15}

& GeoResponder (LLaMa) & 0.167 & 0.945 & 0.341 & 0.178 & 0.449 & 0.27 & 0.058 & 0.82 & 0.287 & 0.7519 & 0.062 & 0.32 & 0.1422 \\
& GeoResponder (Mistral) & 0.221 & 0.979 & 0.407 & 0.254 & 0.502 & 0.348 & 0.025 & 0.86 & 0.531 & 0.7416 & 0.336 & 0.6234 & 0.499 \\
& GeoResponder (Qwen) & 0 & 0 & 0 & 0 & 0 & 0 & 0.159 & 0.75 & 0.215 & 0.7969 & 0.116 & 0.3059 & 0.088 \\
\cmidrule{2-15}

& \textbf{Best GeoResponder} 
& 0.221 & 0.979 & 0.407 
& 0.254 & 0.502 & 0.348 
& 0.025 & 0.86 
& 0.531 
& 0.7416 & 0.336 
& 0.6234 
& 0.499 \\
& \textbf{vs Best Baseline (\%)} 
& \multicolumn{1}{c|}{\cellcolor{green!40}{+146}}  % Hit@1 std
& \multicolumn{1}{c|}{\cellcolor{green!40}{+75}}   % Acc@1km std
& \multicolumn{1}{c|}{\cellcolor{green!40}{+154}}  % MRR std
& \multicolumn{1}{c|}{\cellcolor{green!40}{+217}}  % Hit@1 dir
& \multicolumn{1}{c|}{\cellcolor{green!40}{+113}}  % Acc@1km dir
& \multicolumn{1}{c|}{\cellcolor{green!40}{+188}}  % MRR dir
& \multicolumn{1}{c|}{\cellcolor{green!40}{-95}}   % Dist MAPE ↓ (lower better)
& \multicolumn{1}{c|}{\cellcolor{green!40}{+100}}  % Dir F1
& \multicolumn{1}{c|}{\cellcolor{green!40}{+279}}  % Road Acc
& \multicolumn{1}{c|}{\cellcolor{red!40}{+11}}     % Length MAPE ↓ (worse)
& \multicolumn{1}{c|}{\cellcolor{red!40}{-10}}     % Length A@30%
& \multicolumn{1}{c|}{\cellcolor{green!40}{+102}}  % Speed F1
& \multicolumn{1}{c}{\cellcolor{green!40}{+66}}    % Lanes F1
\\
\midrule

% ------------------ Manila ------------------
\multirow{13}{*}{Manila}
& LLaMa 3.1 8B & 0.004 & 0.077 & 0.01 & 0.001 & 0.011 & 0.003 & 0.52 & 0.1 & 0 & 3.4088 & 0.058 & 0 & 0.0005 \\
& Mistral 7B v0.3 & 0 & 0.013 & 0 & 0.001 & 0.007 & 0.002 & 0.3 & 0.02 & 0 & 7.7808 & 0.01 & 0 & 0.0005 \\
& Qwen3 8B & 0 & 0.011 & 0.001 & 0.001 & 0.008 & 0.002 & 0.503 & 0.04 & 0.01 & 3.3385 & 0.062 & 0 & 0.0388 \\
\cmidrule{2-15}

& GeoPromptFusion (LLaMa) & 0.015 & 0.183 & 0.026 & 0.083 & 0.216 & 0.113 & 0.7701 & 0.09 & 0.029 & 0.7205 & 0.3196 & 0.0814 & 0.1645 \\
& GeoPromptFusion (Mistral) & 0.023 & 0.377 & 0.049 & 0.031 & 0.062 & 0.037 & 22.2 & 0.08 & 0.017 & 0.6809 & 0.2375 & 0.1014 & 0.1486 \\
& GeoPromptFusion (Qweni)& 0.036 & 0.378 & 0.068 & 0.034 & 0.149 & 0.059 & 0.2476 & 0.3 & 0.07 & 0.798 & 0.2335 & 0.1348 & 0.1556 \\
& GeoPromptFusion (GPT-4o mini) & 0.009 & 0.141 & 0.016 & 0.009 & 0.037 & 0.014 & 0.1944 & 0.34 & 0.016 & 0.7158 & 0.339 & 0.0745 & 0.1379 \\
\cmidrule{2-15}

& GeoResponder (LLaMa)   & 0.155 & 0.837 & 0.263 & 0.23 & 0.451 & 0.303 & 0.26 & 0.87 & 0.666 & 0.5303 & 0.343 & 0.6356 & 0.4954 \\
& GeoResponder (Mistral) & 0.257 & 0.977 & 0.403 & 0.28 & 0.478 & 0.352 & 0.02 & 0.95 & 0.785 & 0.3376 & 0.682 & 0.6839 & 0.6481 \\
& GeoResponder (Qwen)    & 0.112 & 0.848 & 0.215 & 0.126 & 0.363 & 0.189 & 0.04 & 0.85 & 0.482 & 0.7147 & 0.186 & 0.4889 & 0.326 \\
\cmidrule{2-15}

& \textbf{Best GeoResponder} 
& 0.257 & 0.977 & 0.403 
& 0.28 & 0.478 & 0.352 
& 0.02 & 0.95 
& 0.785 
& 0.3376 & 0.682 
& 0.6839 
& 0.6481 \\
& \textbf{vs Best Baseline (\%)} 
& \multicolumn{1}{c|}{\cellcolor{green!40}{+614}} 
& \multicolumn{1}{c|}{\cellcolor{green!40}{+158}} 
& \multicolumn{1}{c|}{\cellcolor{green!40}{+493}} 
& \multicolumn{1}{c|}{\cellcolor{green!40}{+237}} 
& \multicolumn{1}{c|}{\cellcolor{green!40}{+121}} 
& \multicolumn{1}{c|}{\cellcolor{green!40}{+212}} 
& \multicolumn{1}{c|}{\cellcolor{green!40}{-90}} 
& \multicolumn{1}{c|}{\cellcolor{green!40}{+179}} 
& \multicolumn{1}{c|}{\cellcolor{green!40}{+1021}} 
& \multicolumn{1}{c|}{\cellcolor{green!40}{-50}} 
& \multicolumn{1}{c|}{\cellcolor{green!40}{+101}} 
& \multicolumn{1}{c|}{\cellcolor{green!40}{+407}} 
& \multicolumn{1}{c}{\cellcolor{green!40}{+294}} \\
\midrule

% ------------------ New York City ------------------
\multirow{13}{*}{New York City}
& LLaMa 3.1 8B   & 0.001 & 0.032 & 0.003 & 0.002 & 0.018 & 0.005 & 0.972 & 0.19  & 0.007 & 1.14   & 0.227 & 0 & 0.04 \\
& Mistral 7B v0.3 & 0.004 & 0.024 & 0.007 & 0.049 & 0.05  & 0.049 & 1.23  & 0.08  & 0.006 & 9      & 0.015 & 0 & 0.0338 \\
& Qwen3 8B    & 0.002 & 0.03  & 0.005 & 0.019 & 0.024 & 0.02  & 0.747 & 0.07  & 0.015 & 2.5578 & 0.122 & 0.01 & 0.104 \\
\cmidrule{2-15}

& GeoPromptFusion (LLaMa)
& 0.01 & 0.2   & 0.028 
& 0.045 & 0.169 & 0.072 
& 0.69  & 0.1 
& 0.0583 
& 0.788 & 0.317 
& 0.1474 & 0.1528 \\
& GeoPromptFusion (Mistral) 
& 0.019 & 0.277 & 0.045 
& 0.012 & 0.07  & 0.023 
& 0.99  & 0.05 
& 0.018 
& 0.76  & 0.323 
& 0.183 & 0.1373 \\
& GeoPromptFusion (Qweni)
& 0.035 & 0.311 & 0.066 
& 0.025 & 0.113 & 0.044 
& 0.86  & 0.36 
& 0.006 
& 0.648 & 0.3584 
& 0.2325 & 0.1557 \\

& GeoPromptFusion (GPT-4o mini) & 0.023 & 0.309 & 0.053 & 0.015 & 0.102 & 0.032 & 0.459 & 0.42 & 0 & 0.9 & 0.29 & 0 & 0.1224 \\
\cmidrule{2-15}

& GeoResponder (LLaMa)   & 0.159 & 0.888 & 0.31 & 0.262 & 0.546 & 0.363 & 0.085 & 0.7 & 0.467 & 0.6696 & 0.377 & 0.3922 & 0.2997 \\
& GeoResponder (Mistral) & 0.265 & 0.979 & 0.463 & 0.329 & 0.612 & 0.437 & 0.03 & 0.91 & 0.621 & 0.4615 & 0.559 & 0.4861 & 0.5029 \\
& GeoResponder (Qwen)    & 0.205 & 0.885 & 0.364 & 0.22 & 0.524 & 0.324 & 0.07 & 0.89 & 0.479 & 0.67 & 0.316 & 0.41 & 0.2501 \\
\cmidrule{2-15}

& \textbf{Best GeoResponder} 
& 0.265 & 0.979 & 0.463 
& 0.329 & 0.612 & 0.437 
& 0.03 & 0.91 
& 0.621 
& 0.4615 & 0.559 
& 0.4861 
& 0.5029 \\

& \textbf{vs Best Baseline (\%)} 
& \multicolumn{1}{c|}{\cellcolor{green!40}{+657}} 
& \multicolumn{1}{c|}{\cellcolor{green!40}{+215}} 
& \multicolumn{1}{c|}{\cellcolor{green!40}{+602}} 
& \multicolumn{1}{c|}{\cellcolor{green!40}{+163}} 
& \multicolumn{1}{c|}{\cellcolor{green!40}{+262}} 
& \multicolumn{1}{c|}{\cellcolor{green!40}{+507}} 
& \multicolumn{1}{c|}{\cellcolor{green!40}{-93}} 
& \multicolumn{1}{c|}{\cellcolor{green!40}{+117}} 
& \multicolumn{1}{c|}{\cellcolor{green!40}{+965}} 
& \multicolumn{1}{c|}{\cellcolor{green!40}{-29}} 
& \multicolumn{1}{c|}{\cellcolor{green!40}{+56}} 
& \multicolumn{1}{c|}{\cellcolor{green!40}{+117}} 
& \multicolumn{1}{c}{\cellcolor{green!40}{+235}} \\

\bottomrule
\end{tabular}
}
\label{tab:all_tasks_full}
\end{table}

Directly prompted LLaMA, Mistral, and Qwen models perform poorly across all settings: nearest-road Hit@1 is near zero, distance MAPE remains high, directional F1 is low, and both road-name and attribute extraction rarely succeed. Geo Prompt Fusion improves these baselines by adding chain-of-thought, OSM context, and geographically proximate few-shot examples, raising Acc@1km to the $0.2$--$0.4$ range and yielding modest gains in segment metadata inference. However, performance remains inconsistent, especially in Paris, Manila, and New York, indicating that inference-time prompting alone is not enough.

GeoResponder closes this gap substantially. Because its curriculum explicitly grounds entities in the coordinate manifold (Layer~1), internalizes geometric operators (Layer~2), and practices constraint-based retrieval (Layer~3), it achieves robust free-form reasoning without specialized prompts. Across all cities, the strongest GeoResponder variant improves nearest-road Hit@1 by +146\%--+657\%, directional Hit@1 by +163\%--+237\%, and reduces distance MAPE by 80\%--95\%. Road-name accuracy rises from below $0.16$ in all baselines to $0.47$--$0.79$, while speed and lane F1-scores reach $0.39$--$0.68$ and $0.43$--$0.65$. Gains are largest in Manila and New York, where large and irregular road networks impose significant generalization challenges. Among backbones, Mistral performs best overall, with LLaMA close behind; Qwen is strong primarily in Manila and New York. The main remaining difficulty is segment-length estimation in Paris, reflecting the complexity of dense European urban geometry.
%}

\subsection{Out-of-distribution MCQ tasks}
\begin{figure}[t]
    \centering
    \begin{subfigure}{0.48\linewidth}
        \centering
        \includegraphics[width=\linewidth]{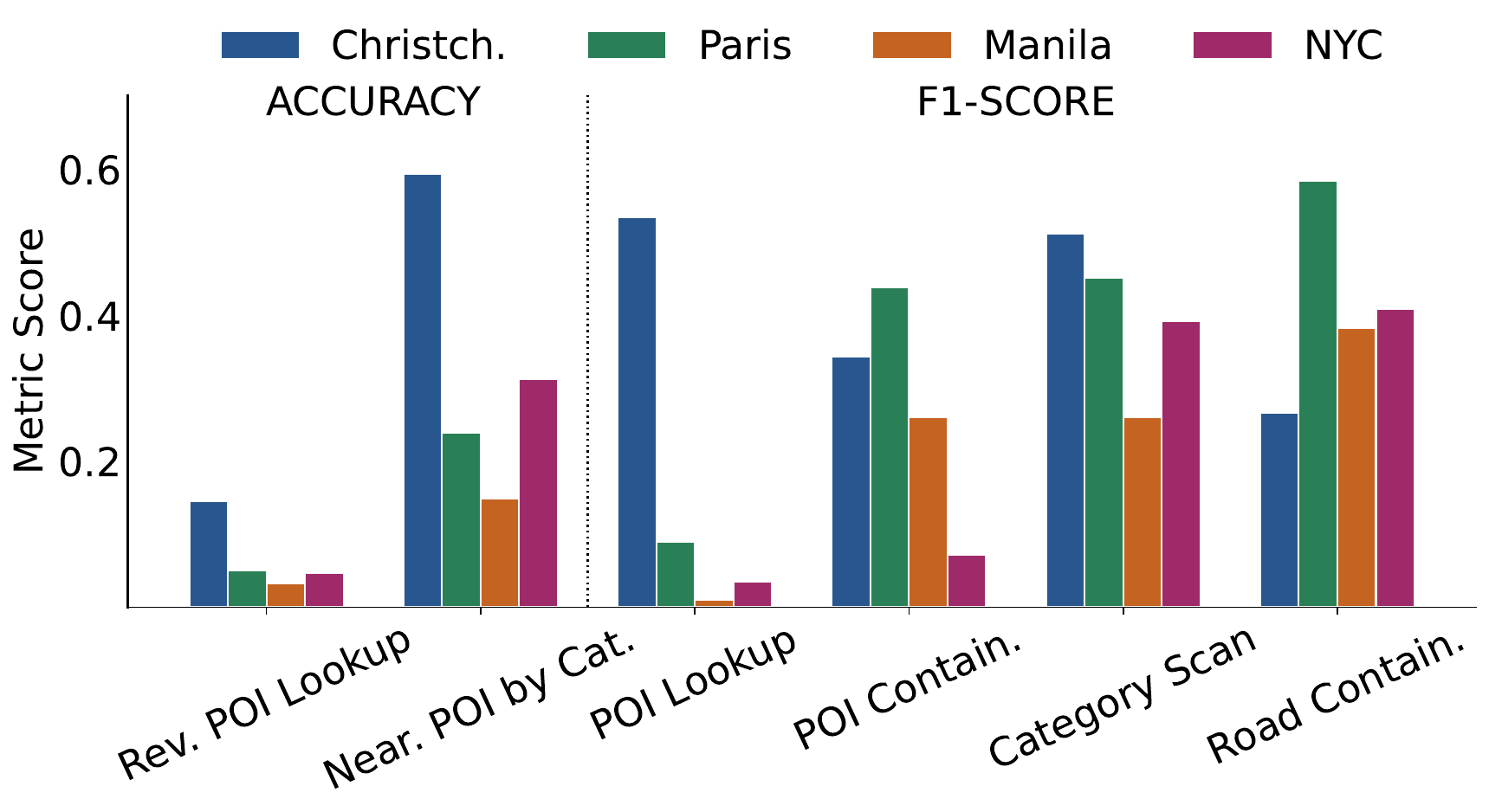}
        \caption{In-dist. non-MCQ evaluation.}
        \label{fig:in_distribution_non_mcq_bars}
    \end{subfigure}
    \hfill
    \begin{subfigure}{0.48\linewidth}
        \centering
        \includegraphics[width=\linewidth]{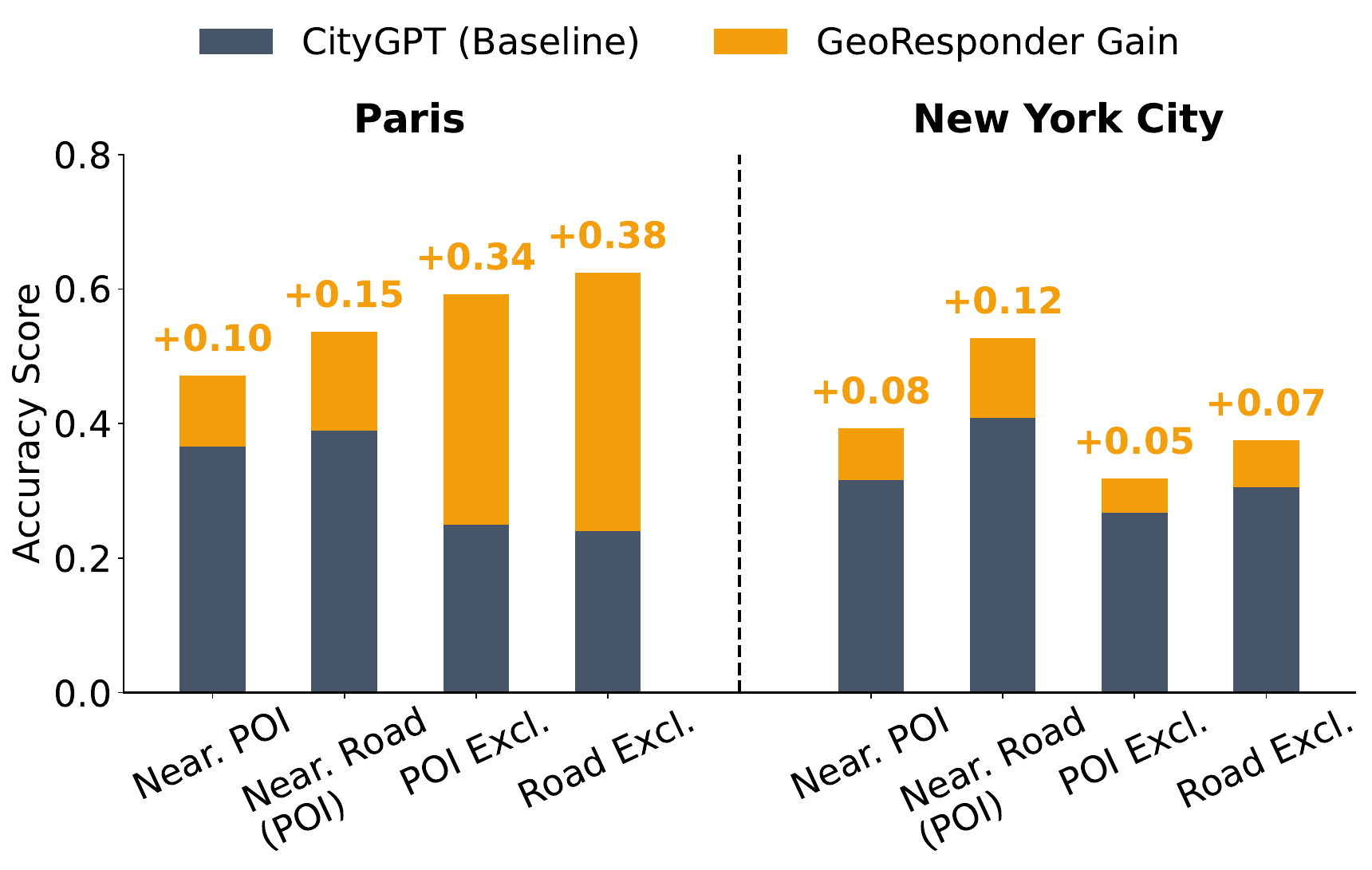}
        \caption{OOD disaster tasks evaluation}
        \label{fig:disaster_ood_tasks}
    \end{subfigure}

\end{figure}

%{\color{blue}
Figure~\ref{fig:disaster_ood_tasks} reports accuracy on the out-of-distribution MCQ tasks in Paris and New York City. These evaluations probe whether models can transfer the grounded and geometric competencies learned to task formats not seen during training.

GeoResponder (Mistral) achieves consistent gains across all OOD tasks, with the largest improvements on POI Exclusion and Road Exclusion. Both tasks require multi-step spatial filtering—identifying which entities lie \emph{outside} a specified region—thus stressing the model’s ability to combine zonal scanning with fine-grained coordinate reasoning. The strong performance suggests that GeoResponder’s constraint-aware retrieval layer generalizes effectively even when the task logic differs from the atomic representations used during training.

CityGPT exhibits notably lower accuracy on these exclusion-based tasks, especially those involving precise coordinate comparisons. Its performance trends mirror the challenges observed in the in-distribution evaluation, indicating that it generalizes less reliably to spatial structures that deviate from its training format. By contrast, GeoResponder maintains robust accuracy across both cities and task types, highlighting its stronger capacity to transfer spatial reasoning beyond the distributions and representations it was explicitly trained on.
%}

\subsection{In-Distribution Non-MCQ Tasks}
%{\color{blue}
Figure~\ref{fig:in_distribution_non_mcq_bars} reports GeoResponder (Mistral) performance on non-MCQ tasks across all four cities. Reverse POI Lookup emerges as the most challenging task: mapping from the continuous space to a discrete entity is inherently difficult and becomes increasingly ambiguous as POI density grows. Similarly, POI Lookup accuracy degrades in dense cities like Manila and NYC, confirming that large, crowded candidate spaces impede retrieval. %A similar trend appears in POI Lookup, where accuracy remains high in Christchurch but drops sharply in Manila and New York City, confirming that lookup-style tasks become harder when the candidate space is large and spatially crowded.
Performance on the remaining tasks is more consistent. POI Containment and Category Scan exhibit moderate variation across cities, with lower accuracy in denser environments where POI clusters frequently overlap. Road Containment is the most stable task overall, suggesting that reasoning over polygon–polyline relationships is less affected by city scale and density than point-based retrieval. These results indicate that non-MCQ accuracy is influenced primarily by density and urban complexity, and that GeoResponder generalizes more reliably for region-based reasoning than for fine-grained point-to-point localization.
%}

% now we will talk about In distribution non mcq tasks. the figure shows the results of georesponder mistral across the four cities.
% we can see that rev poi lookup is the most difficult task, which is mapping from the entity space to the continous space. similarly  the model struggles with poi lookup except for chnz where it does exceedigly hjigh, most likely meaning that the the difficulty of this task is directly linked to the POI density of the city (also talk a bit about the other tasks)

\input{case_studies}

%% file: case_studies.tex
\subsection{Qualitative Assessment}

\begin{figure}[h]
\centering
\includegraphics[height=3.5cm]{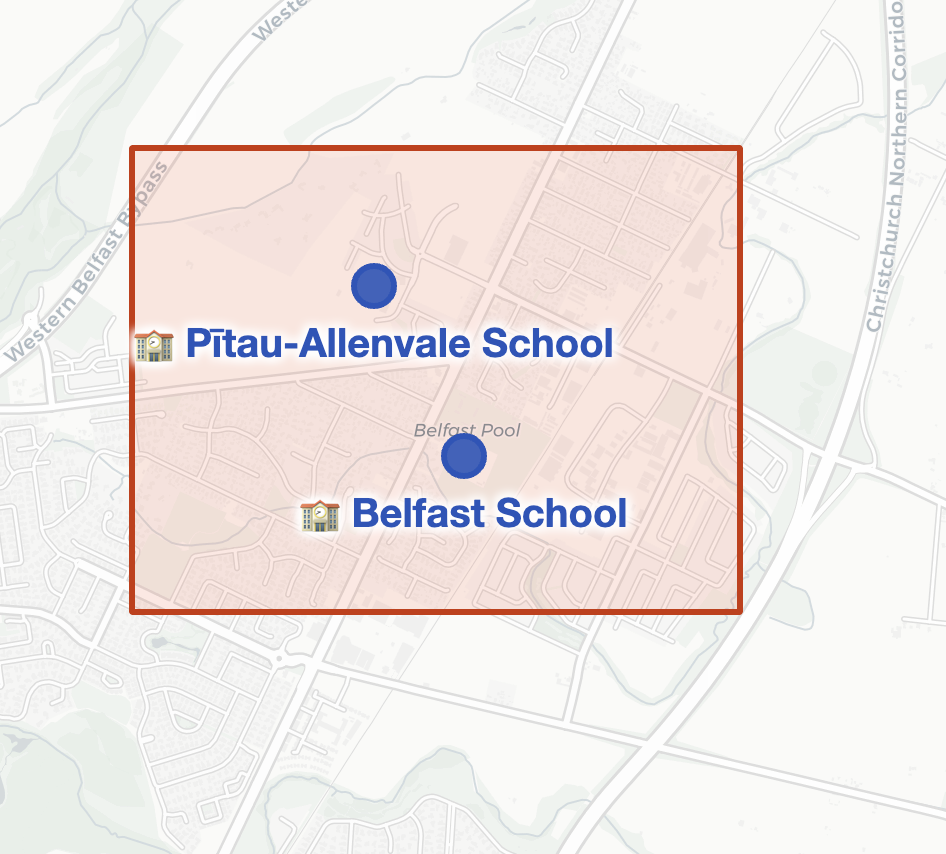}
\hfill
\includegraphics[height=3.5cm]{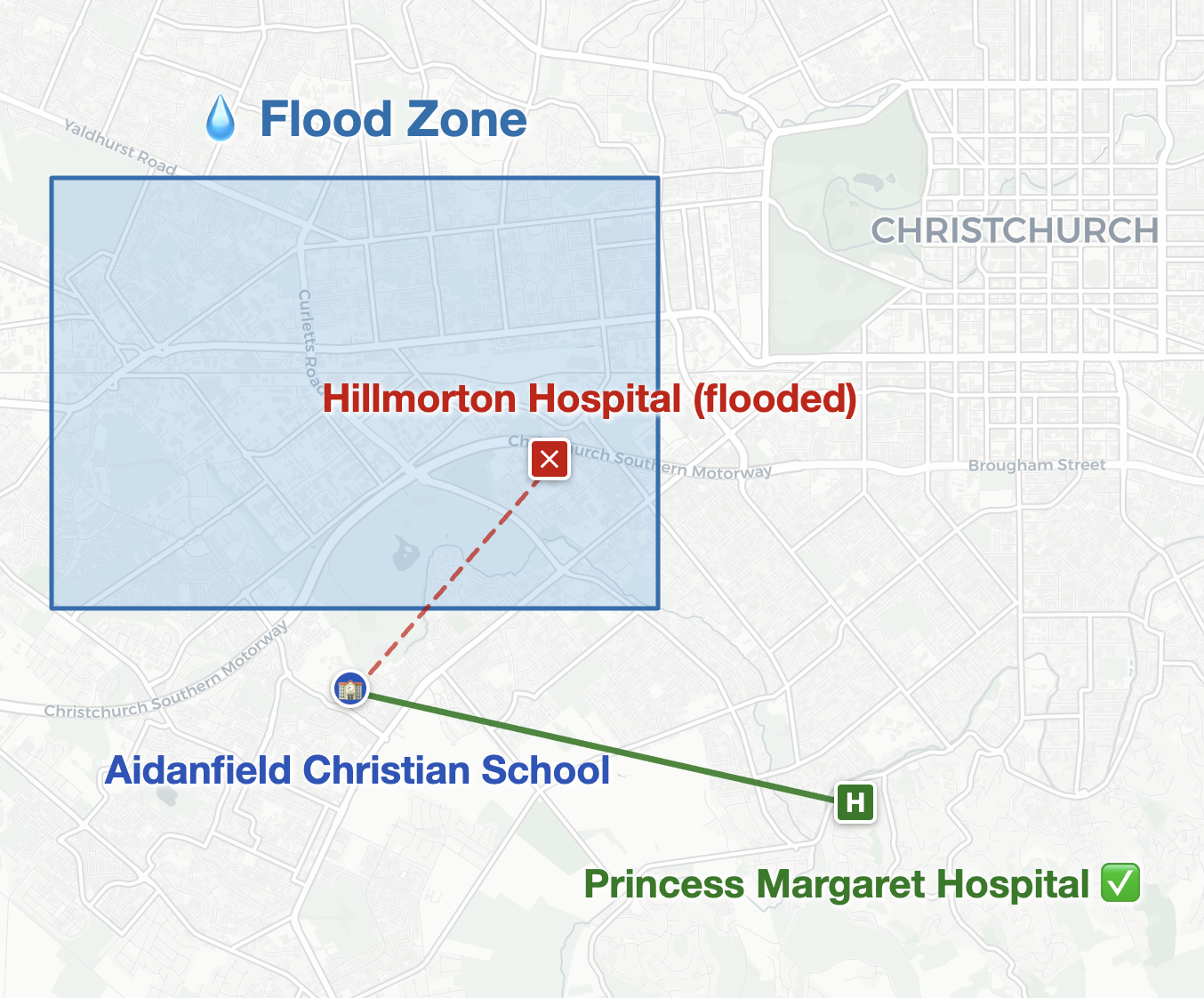}
\hfill
\includegraphics[height=3.5cm]{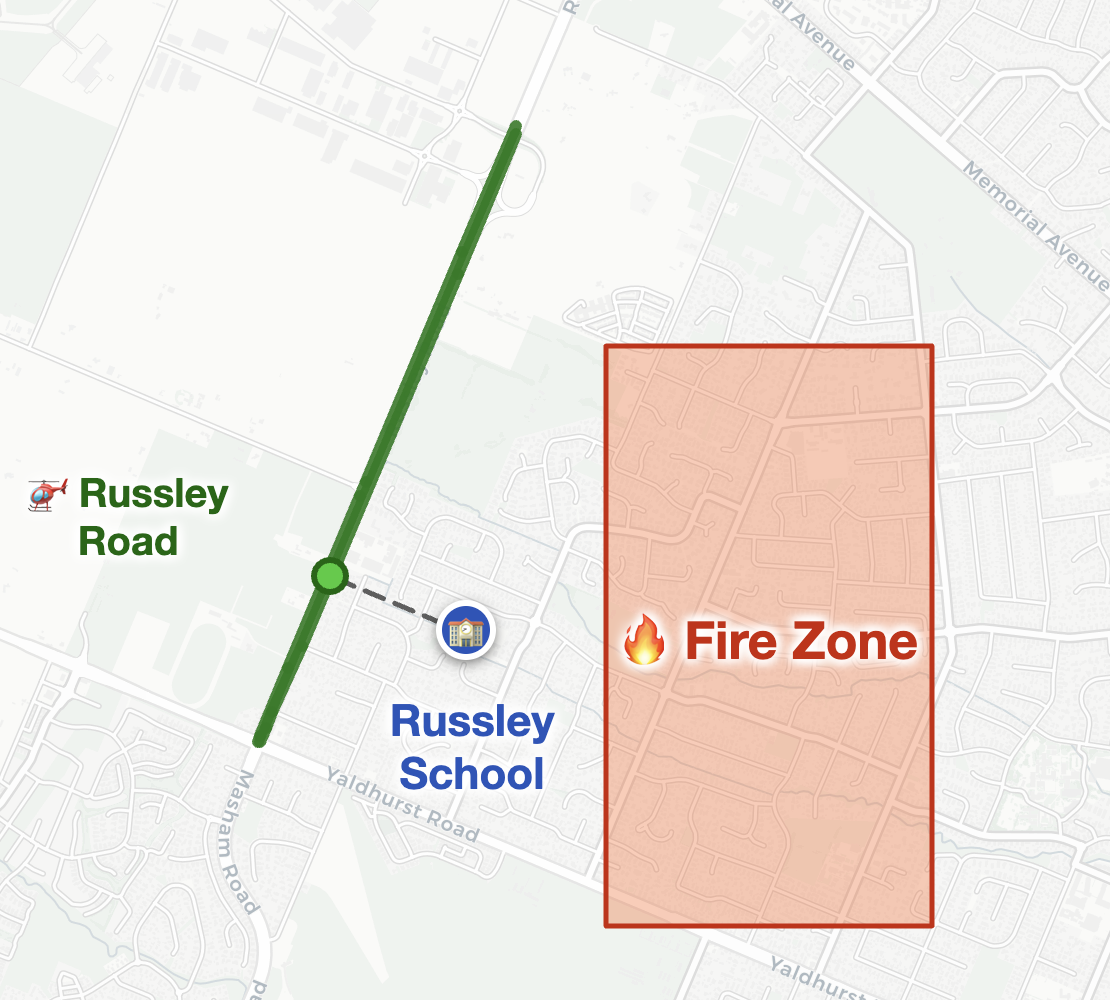}
\caption{Visual representation of GeoResponder's solutions for disaster scenarios.}
\label{fig:case_studies}
\end{figure}

Fig.~\ref{fig:case_studies} presents three examples illustrating how GeoResponder handles realistic disaster-response queries requiring multiple geospatial reasoning capabilities.

\textbf{Explosion scenario.}
\textit{Query:} ``There was an explotion heard near {\color{blue}Northern Belfast}. What nearby {\color{blue}schools} could be used as temporary shelters?''
Answering this query requires identifying candidate schools near the affected region and retrieving viable nearby options that could serve as shelters. GeoResponder returns several schools in the surrounding area that can serve as temporary shelters.

\textbf{Flood scenario.}
\textit{Query:} ``I am at {\color{blue}Aidanfield Christian School} and a flood happened around {\color{red}Curletts Road}
% in bbox {\color{red}(-43.557506, 172.543373, -43.527393, 172.631950)}. 
What is the closest safe hospital to where I am?''
Answering this query requires grounding the user location, identifying candidate hospitals, filtering those inside the flooded region, and computing distances to determine the nearest safe option. GeoResponder returns {\color{darkgreen}Princess Margaret Hospital}, 5\,km away, as the closest hospital outside the affected area.

\textbf{Fire scenario.}
\textit{Query:} ``There is a {\color{red}fire} raging to the {\color{red}east} of {\color{blue}Russley School}. A rescue helicopter is looking to land near the school. What is the nearest road of type highway to the west of the school?''
Solving this query requires directional reasoning, road-type filtering, and nearest-neighbor search over the road network. GeoResponder identifies {\color{darkgreen}Russley Road}, approximately 100\,m away.

Unlike Figure~\ref{fig:qualitative_hospital}, we do not display the responses of baseline models (CityGPT and 4o-mini) because their outputs are not viable; in these scenarios they fail to return any correct schools or roads in Christchurch. These examples illustrate how GeoResponder composes capabilities learned through our geospatial representations, including spatial grounding, directional reasoning, constraint filtering, and nearest-entity retrieval, to answer complex disaster-response queries.

%% file: ablation.tex
\section{Ablation}

\begin{figure}[t]
    \centering
    % Width can be adjusted (0.75\textwidth is common)
    \includegraphics[width=0.7\textwidth]{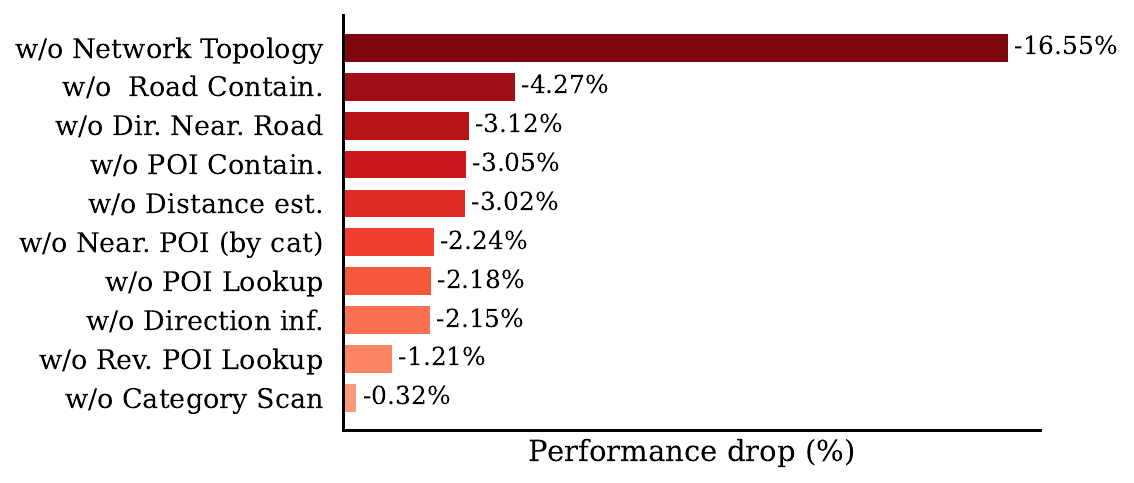}

    \caption{
        Ablation study showing the average performance drop (on the average of all evaluation tasks) when individual representations are removed. %Ablation study: drop in averaged performance across all tasks when removing individual data representations. 
        %Each bar shows the reduction in the unified metric when its corresponding representation is ablated.
    }

    \label{fig:ablation_drop}
\end{figure}

Figure~\ref{fig:ablation_drop} summarizes the contribution of each geospatial representation by removing one at a time and measuring the resulting drop in downstream performance. Every ablation causes a decline, confirming that all representations contribute to geospatial capability. The largest drop occurs when removing \emph{network topology} ($\approx$16.6\%), which is expected since it forms the largest portion of the training data (Table~\ref{tab:city_stats_training}) and encodes structural constraints of the road graph. Without it, the model loses critical connectivity cues for routing, adjacency reasoning, and multi-hop inference. Other representations produce notable drops: removing \emph{Road Containment}, \emph{Directional Nearest Road}, or \emph{POI Containment} reduces performance by 3--4\%, highlighting their role in regional and directional reasoning. Point-based tasks such as POI Lookup, Reverse POI Lookup, and Direction Inference yield smaller but consistent declines, while Category Scan, the least impactful ablation, produces a measurable drop. Overall, the results show that GeoResponder benefits from the complementary structure of representations: strong performance emerges from combining topological, directional, containment, and retrieval-based signals within a unified geospatial training framework.

% ----------- Figure 5: Ablation plot

%% file: conclusion.tex
\section{Conclusion}

We addressed the fundamental deficit of geospatial reasoning in Large Language Models by introducing GeoResponder, a framework that decouples spatial intelligence into a scaffolded curriculum of grounding, reasoning, and constraint-aware retrieval. Our empirical evaluation across heterogeneous urban environments demonstrates that shifting from latent textual correlation to structured topological supervision significantly improves performance. Crucially, our results establish that internalized spatial representations offer a robust, low-latency complement to brittle agentic workflows, enabling models to resolve complex, constraint-heavy optimization problems directly within their neural weights. Moving forward, we aim to extend this paradigm by integrating multi-modal inputs, such as satellite imagery and real-time sensor streams, paving the way for fully autonomous and resilient geospatial decision support systems.